\PassOptionsToPackage{table}{xcolor}
\documentclass[10pt,twocolumn,letterpaper]{article}
\usepackage[pagenumbers]{iccv}

\renewcommand{\ie}{\textit{i}.\textit{e}., }
\renewcommand{\eg}{\textit{e}.\textit{g}., }

\usepackage{multirow}
\usepackage{xcolor}
\usepackage{graphicx}
\usepackage{float}
\usepackage{duckuments}
\usepackage[table]{xcolor}
\usepackage{soul}
\usepackage{booktabs}
\usepackage{pifont}

\newcommand{\cmark}{\ding{51}}
\newcommand{\xmark}{\ding{55}}
\newcommand{\methodname}{FontAdapter\xspace}
\newcommand{\methodnameshort}{FontAdapter}

\newcommand\blfootnote[1]{
  \begingroup
  \renewcommand\thefootnote{}\footnote{#1}
  \addtocounter{footnote}{-1}
  \endgroup
}

\newcommand{\scolor}{\scolor{blue}}

\definecolor{iccvblue}{rgb}{0.21,0.49,0.74}
\usepackage[pagebackref,breaklinks,colorlinks,allcolors=iccvblue]{hyperref}

\title{\methodname: Instant Font Adaptation in Visual Text Generation}

\author{
    Myungkyu Koo$^{1}$ \hspace{6pt}
    Subin Kim$^{1}$ \hspace{6pt}
    Sangkyung Kwak$^{2, \dagger}$ \hspace{6pt}
    Jaehyun Nam$^{1}$ \hspace{6pt}
    Seojin Kim$^{3, \dagger}$ \hspace{6pt}
    Jinwoo Shin$^{1}$ \\
    $^{1}$KAIST \hspace{10pt}
    $^{2}$General Robotics \hspace{10pt}
    $^{3}$Seoul National University \\
    \tt\small jameskoo0503@kaist.ac.kr
}

\begin{document}
\maketitle

\begin{abstract}

\blfootnote{
Project page: \href{https://fontadapter.github.io/}{https://fontadapter.github.io/}
\hspace{6pt}$^{\dagger}$ Work done at KAIST.
}
Text-to-image diffusion models have significantly improved the seamless integration of visual text into diverse image contexts.
Recent approaches further improve control over font styles through fine-tuning with predefined font dictionaries.
However, adapting unseen fonts outside the preset is computationally expensive, often requiring tens of minutes, making real-time customization impractical.
In this paper, we present \methodname, a framework that enables visual text generation in unseen fonts within seconds, conditioned on a reference glyph image.
To this end, we find that direct training on font datasets fails to capture nuanced font attributes, limiting generalization to new glyphs.
To overcome this, we propose a two-stage curriculum learning approach: \methodname first learns to extract font attributes from isolated glyphs and then integrates these styles into diverse natural backgrounds.
To support this two-stage training scheme, we construct synthetic datasets tailored to each stage, leveraging large-scale online fonts effectively.
Experiments demonstrate that \methodname enables high-quality, robust font customization across unseen fonts without additional fine-tuning during inference.
Furthermore, it supports visual text editing, font style blending, and cross-lingual font transfer, positioning \methodname as a versatile framework for font customization tasks.

\end{abstract}
\section{Introduction}

Diffusion-based text-to-image (T2I) generative models~\citep{rombach2022high, saharia2022photorealistic, deepfloydif} have shown remarkable capabilities in generating realistic images with high fidelity to user prompts.
Notably, recent models such as Stable Diffusion 3~\citep[SD3;][]{esser2024scaling}, DaLLE$\cdot$3~\citep{betker2023improving, dalle3}, and FLUX.1~\citep{blackforestlab2024flux}, have achieved significant progress in visual text generation, \ie accurately rendering glyphs within diverse image contexts.

Several works~\citep{ji2023improving, liu2025glyph, liu2024glyph} enhance these capabilities to control font styles through fine-tuning with predefined font dictionaries.
However customizing unseen fonts using these approaches incurs high computational costs, limiting their usage when handling numerous unique fonts: \eg fine-tuning SD3 for font customization with LoRA~\citep{hu2021lora} takes about 40 minutes on 4 RTX 3090 GPUs.

\begin{figure*}[t]
    \centering
    \captionsetup{type=figure}
    \includegraphics[width=\textwidth]{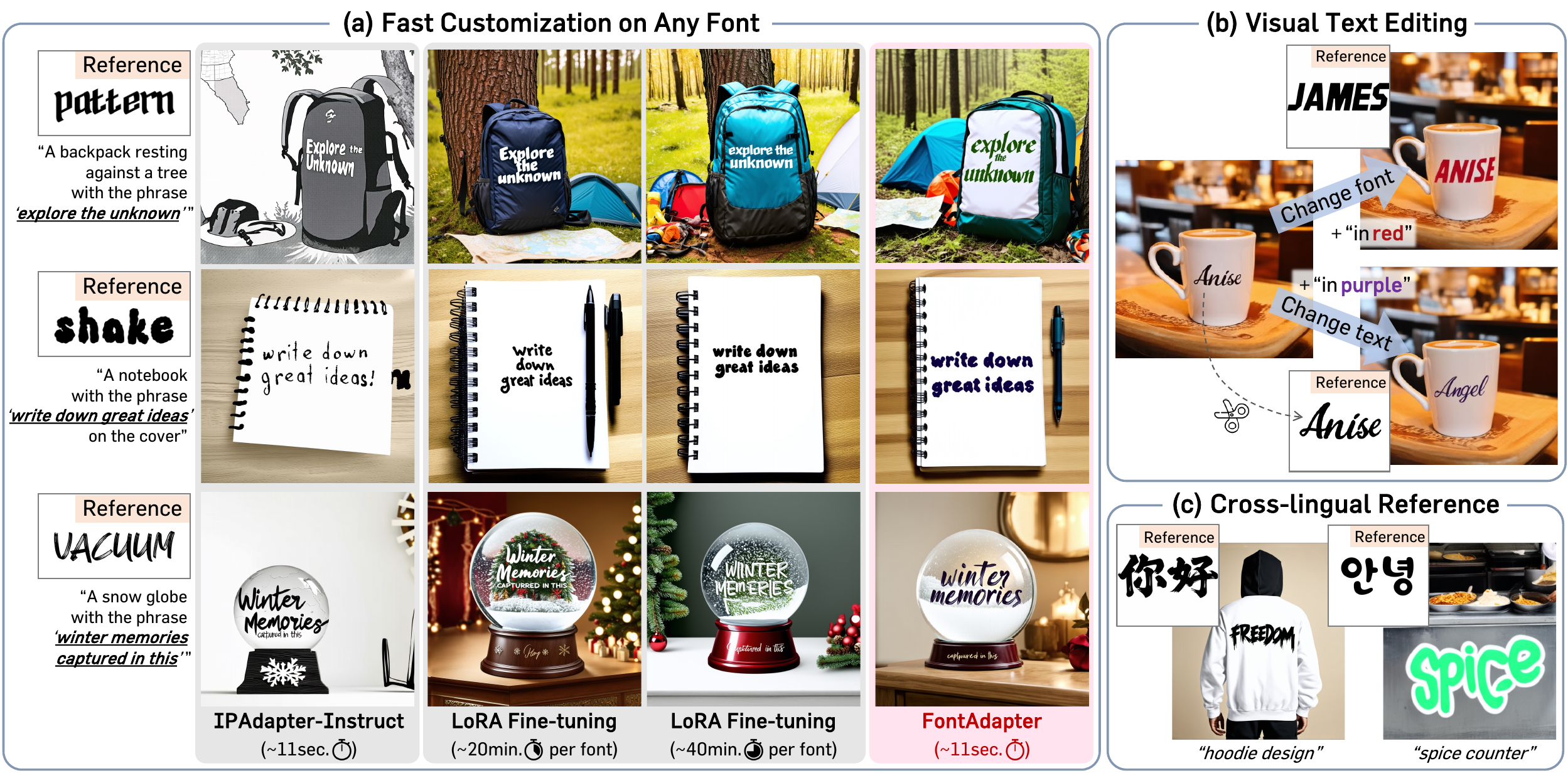}
    \vspace{-7mm}
    \captionof{figure}{\textbf{Font customization with \methodname.}
        In this paper, we present \methodname, a framework for font customization that instantly adapts a wide range of font styles across diverse image contexts.
        (a) shows that \methodname enables fast font customization, while successfully capturing both the nuances and intricate details of unique font styles.
        (b) and (c) highlight the versatility of \methodname, including controlled editing of visual text and the adaptation of font styles from diverse languages.
    }
    \label{fig:teaser}
    \vspace{-3mm}
\end{figure*}

Meanwhile, recent approaches~\citep {wei2023elite, shi2024instantbooth, ye2023ip} enable customization of specific subjects or styles when generating images conditioned on a reference image, without additional fine-tuning.
Although these methods demonstrate strong generalizability for general styles (\eg composition, painting style), they fall short when adapting font styles (see Figure~\ref{fig:teaser}(a)).
Specifically, given a glyph image on a white background, they struggle to properly extract font attributes, often yielding overly simplistic and monotonous results.

In this paper, we propose \methodname, a framework for instant image-based font customization in visual text generation.
\methodname eliminates the need for time-consuming fine-tuning while precisely adapting any font style to visual text in realistic backgrounds.
To achieve this, we introduce a novel two-stage curriculum training scheme, as we observe that na\"ive training on font-specific datasets struggles to capture detailed font features.
\methodname first learns to extract font attributes from a given image without interference from other visual elements.
Once it successfully extracts and applies these attributes to new glyphs, it then focuses on rendering them within complex backgrounds.

Effectively implementing this two-stage curriculum requires constructing datasets tailored to each stage.
However, collecting in-the-wild visual text in a consistent font is challenging.
To address this, we introduce scalable, synthetic, font-specific datasets by leveraging diverse online fonts.
For the first stage, we construct a \textit{text-only} dataset with black glyphs on white backgrounds, allowing \methodname to extract glyph details without background distractions.
For the second stage, we introduce a \textit{scene-text} dataset, where colorized glyphs are naturally integrated into realistic backgrounds.
By aligning stage-specific datasets with our two-stage curriculum, \methodname effectively learns font attributes and adapts them to complex backgrounds, enabling high-quality visual text generation.

We comprehensively evaluate font similarity by comparing the generated visual text to its ground-truth glyph, introducing a new evaluation pipeline that examines multiple aspects.
\methodname significantly outperforms baseline methods while mitigating text accuracy degradation and prompt misalignment---common issues in conditioning-based approaches.
Furthermore, we demonstrate diverse visual font applications, including long text renderings, visual text editing, cross-lingual font transfer, and font blending.

\vspace{0.05in}
\noindent\textbf{Contributions.} Our contributions are as follows:
\begin{itemize}
    \item We propose \methodname, a framework enabling precise visual text customization in just 11 seconds on a single RTX 3090 GPU---substantially reducing computational cost compared to fine-tuning-based methods.

    \item \textit{Two-stage curriculum training scheme}: We divide font customization ability into two stages---first extracting font attributes, then adapting them to diverse image contexts---addressing limitations of na\"ive training.

    \item \textit{Synthetic dataset construction}: We construct stage-specific datasets by leveraging large-scale online fonts.
    
    \item \textit{Comprehensive evaluation}: We evaluate font customization in three key aspects: font similarity, accuracy, and alignment, demonstrating \methodname's consistent superiority over baseline methods.

    \item \textit{Broad applicability}: \methodname supports various applications, including visual text editing, cross-lingual font adaptation, and font style blending, establishing it as a versatile framework for visual text customization tasks.
\end{itemize}
\begin{figure*}[ht]
    \centering\small
    \includegraphics[width=0.95\textwidth]{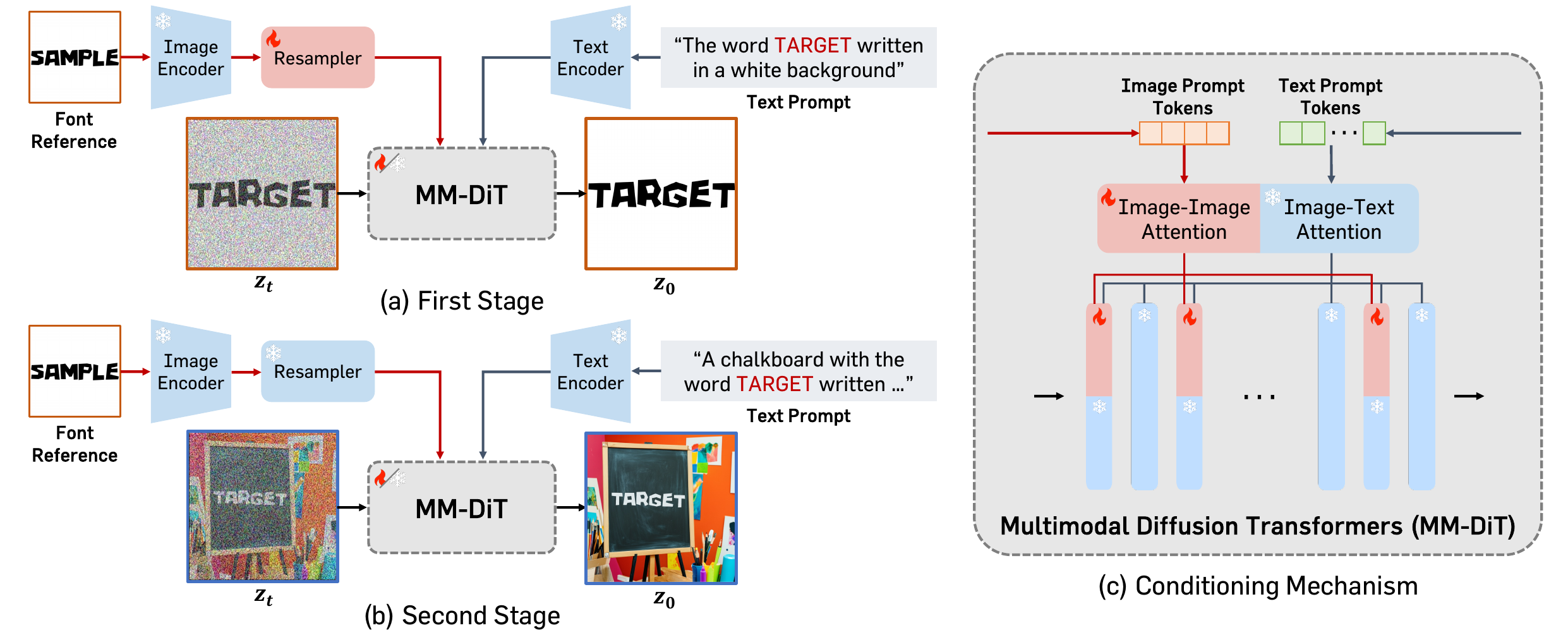}
    \vspace{-2mm}
    \caption{\textbf{Overview.}
        (a) In the first stage, the model is trained to generate a \textit{text-only} image, allowing the model to effectively understand font attributes.
        (b) In the second stage, the model is trained to generate a \textit{scene-text} image, allowing the model to learn the incorporation of visual text and the background. The Resampler module is frozen in this stage to preserve the font attributes previously learned.
        (c) In the MM-DiT, image prompt tokens extracted from the font image embeddings are fed into image-image attention layers in parallel with the original image-text attentions.
        In both stages, training is applied only to these added image-image attention layers.
    }
    \vspace{-2mm}
    \label{fig:overview}
\end{figure*}

\section{Preliminaries}
\label{method:preliminary}

\vspace{0.05in}
\noindent\textbf{T2I Diffusion Models.} 
Text-to-image (T2I) diffusion models generate images from textual prompts via iterative denoising or flow-based transformations.
In particular, Stable Diffusion 3~\citep[SD3;][]{esser2024scaling} adopts a rectified flow~\citep{liu2022flow} framework, which models a velocity field transporting noise vectors $\mathbf{z}_0$ to image latents $\mathbf{z}_1$ along linear interpolation paths.
SD3 operates in a latent space defined by a variational autoencoder~\citep[VAE;][]{van2017neural} and conditions generation on text prompts using a pretrained encoder (\eg CLIP~\citep{radford2021learning} text encoder).
During training, SD3 learns the velocity field $\boldsymbol{v}_\theta$ by minimizing the conditional flow matching (CFM) objective:
\begin{equation}
    \mathcal{L}_{\tt CFM} = \mathbb{E}_{t, \mathbf{z}_0} \left[ \| \boldsymbol{v} - \boldsymbol{v}_\theta(\mathbf{z}_t, t, \mathbf{c}_\mathbf{t}) \|^2_2 \right],
\end{equation}
where the target velocity is defined as $\boldsymbol{v} = \mathbf{z}_1 - \mathbf{z}_0$, and $\boldsymbol{v}_\theta$ denotes the predicted velocity given the interpolated latent $\mathbf{z}_t = (1-t)\mathbf{z}_0 + t\mathbf{z}_1$, timestep $t$, and text condition $\mathbf{c}_\mathbf{t}$.

\vspace{0.05in}
\noindent\textbf{Image Prompt Adapter.}
\citet{ye2023ip} propose the Image Prompt Adapter (IP-Adapter), enabling image input as additional conditioning.
This method extends pretrained T2I models by incorporating a visual encoder (\eg CLIP~\citep{radford2021learning} image encoder) to extract features from a reference image $\mathbf{c}_\mathbf{i}$.
These features are then projected via a linear layer or the Resampler module~\citep{alayrac2022flamingo} into an added attention layer parallel to the text-conditioning layer.
During training, original parameters of the pretrained model remain frozen: only newly added projection and attention layers are fine-tuned, integrating both textual and visual cues.
Due to its versatility, various following works have extended IP-Adapter by adding functionalities, including IPAdapter-Instruct~\citep{rowles2024ipadapter} or domain-specific models~\citep{choi2024improving}.
\section{Method}

In this section, we first define our problem setup on instant font customization and highlight its key challenges (Section~\ref{method:setup}).
We introduce the synthetic datasets (Section~\ref{method:synthetic_data}) carefully designed for our two-stage training.
We then detail our curriculum training scheme (Section~\ref{method:two_stage}), followed by an additional domain alignment stage to reduce domain gaps between synthetic and real-world images (Section~\ref{method:domain_alignment}).

\subsection{Problem Setup}
\label{method:setup}

\begin{figure}[!t]
    \centering\small
    \includegraphics[width=0.9\linewidth]{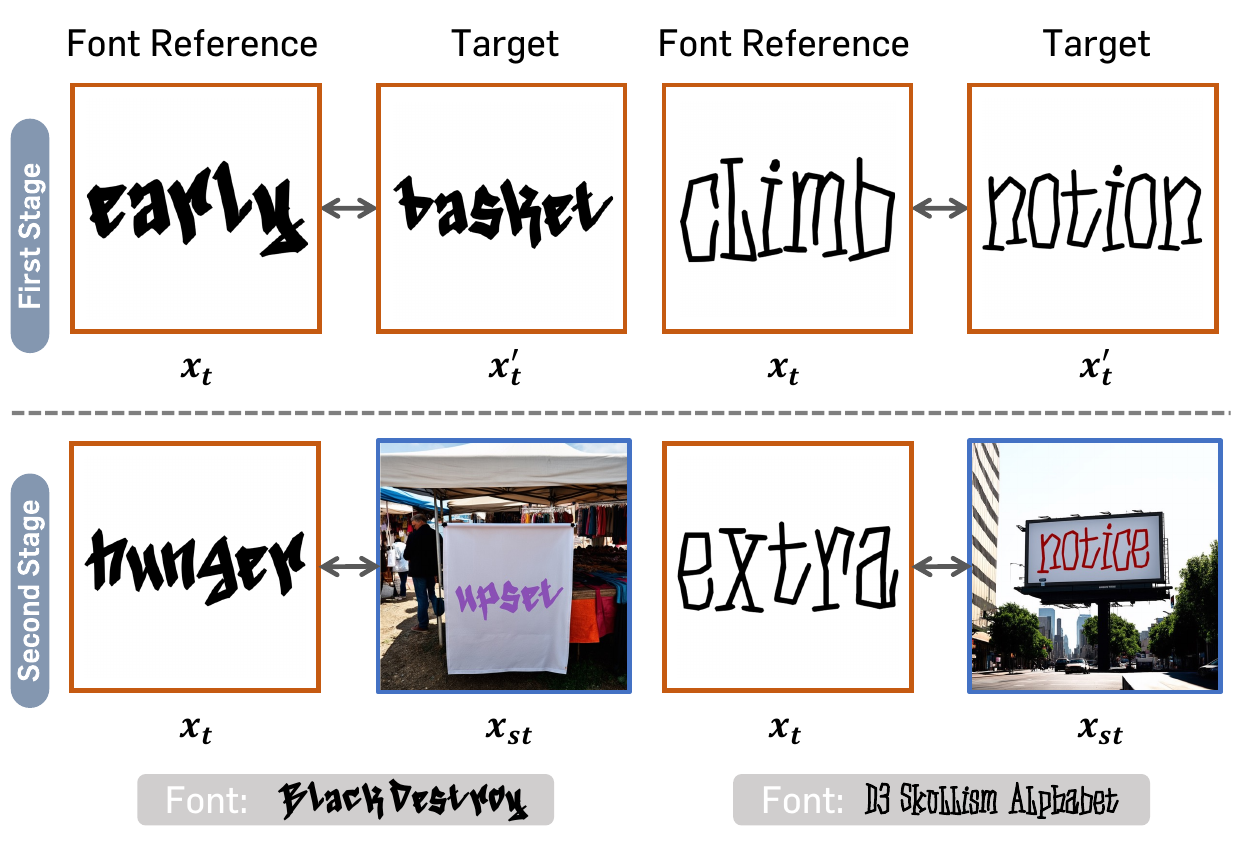}
    \vspace{-2mm}
    \caption{\textbf{Examples of synthetic paired datasets.}
        \sethlcolor{Orange!15} \hl{\textit{Text-only images}} are paired with each other in the first stage (top row) and are paired with \sethlcolor{RoyalBlue!15} \hl{\textit{scene-text images}} in the second stage (bottom row).
        Note that each pair share same font styles while differs in text.
    }
    \label{fig:dataset}
    \vspace{-3mm}
\end{figure}

Our goal is to generate a \textit{visual text image with a customized font style} $\mathbf{x} \in \mathbb{R}^{H \times W \times 3}$ from a conditional distribution $p(\mathbf{x}|\mathbf{c}_\mathbf{t}, \mathbf{c}_\mathbf{f})$, where $\mathbf{c}_\mathbf{t}$ is a text prompt for the visual text image to be generated, and $\mathbf{c}_\mathbf{f}$ is a font reference image for font customization. 
Here, we use $\mathbf{c}_\mathbf{f}$ as black glyphs on white backgrounds (see font reference examples in Figure~\ref{fig:dataset}) because off-the-shelf text segmentation models (\eg TexRNet~\citep{xu2021rethinking}) make it easy to obtain such reference examples from the in-the-wild visual text images.

In this paper, we focus on \textit{instant} font customization, \ie building a \textit{single} model $p_{\mathrm{\theta}}(\mathbf{x}|\mathbf{c}_\mathbf{t}, \mathbf{c}_\mathbf{f})$ capable of generating visual text images with a customized font style for \textit{any} choice of $\mathbf{c}_\mathbf{t}$ and $\mathbf{c}_\mathbf{f}$.
Since font-specific fine-tuning requires significant training costs and storage to build separate models for each font style, our setup, which eliminates this tuning, is critical for practical font customization.
We draw motivation from the recent conditioning-based approaches of IP-Adapter~\citep{ye2023ip} in customized image generation.
Similar to IP-Adapter, we train our model using the following flow matching loss:
$
    \mathcal{L} := \mathbb{E}_{t, \mathbf{z}_0} \left[ \|\boldsymbol{v} - \boldsymbol{v}_\theta(\mathbf{z}_t, t, \mathbf{c}_\mathbf{t}, \mathbf{c}_\mathbf{i})\|^2_2 \right],
$
where $\mathbf{c}_\mathbf{t}$ and $\mathbf{c}_\mathbf{i}$ represent token embeddings of the text prompt and font reference image, respectively.

\vspace{0.05in}
\noindent\textbf{Key Challenges of Instant Font Customization.}
However, we note that both the direct application of IP-Adapter and further fine-tuning on font-specific datasets fall short in font customization.
The direct application fails to disentangle font styles from glyphs and backgrounds in the reference image $\mathbf{c}_\mathbf{f}$, yielding visual text with simplistic backgrounds and poor adaptation of nuanced font styles to new glyphs (see Figure~\ref{fig:qualitative}, middle row).
Even with further fine-tuning, fine details from the reference images are inadequately captured (\eg the bent edge of the reference `T' is not preserved in Figure~\ref{fig:freeze_resampler}).
To address this, we propose a curriculum learning approach that divides font customization into two stages: first, effectively disentangling and applying font attributes, and then integrating them into realistic backgrounds according to the given prompts.

\subsection{Dataset Construction}
\label{method:synthetic_data}

We construct font-specific datasets where each target image is paired with its corresponding font reference (see Figure~\ref{fig:dataset}).
As discussed in Section~\ref{method:setup}, we divide the training scheme into two stages to effectively learn both font attribute extraction from a reference image and style adaptation to diverse image contexts.
To achieve this, we generate two categories of images: \textit{text-only} and \textit{scene-text}, leveraging a large-scale online source of diverse font styles.

\vspace{0.05in}
\noindent\textbf{\textit{Text-only} Images and \textit{Scene-text} Images.}
To generate \textit{text-only} images, we render black glyphs centered on white backgrounds.
For \textit{scene-text} images, where colorized glyphs are embedded into realistic scenes, we first create images with designated empty regions for text placement.
Then, we employ Stable Diffusion 3~\citep[SD3;][]{esser2024scaling} with prompts (\eg \textit{``An empty price board at a vegetable stall in a bustling farmers market''}) generated by GPT-4o~\citep{OpenAI2024ChatGPT4o}.
These empty regions are manually labeled with four corner coordinates to ensure precise text placement.
Next, we render random new words in the same font onto the backgrounds, warping them to fit within the labeled boxes while preserving each glyph’s aspect ratio.
To prevent overfitting in colorization, glyphs in scene-text images are randomly colored.
See Appendix~\ref{suppl:datasets} for more details and examples.

\vspace{0.05in}
\noindent\textbf{Constructing Paired Datasets for Each Stage.}
To utilize two-stage training, we synthesize paired datasets for each stage, comprising two categories of images.
Each pair maintains the same font style but differs in glyphs. 
Specifically, we construct: (1) $\mathcal{D}_\mathbf{t}$: A dataset of multiple pairs of \textit{text-only} images, and (2) $\mathcal{D}_\mathbf{st}$: A dataset of \textit{scene-text} images paired with corresponding \textit{text-only} images (see Figure~\ref{fig:dataset}).
Crucially, varying the text content within each pair ensures that \methodname learns glyph-agnostic font styles rather than replicating reference glyphs, enabling adaptation to unseen glyphs at inference time (see Section~\ref{sec:ablations} for an analysis).
Interestingly, this also allows \methodname to support cross-lingual customization (see Section~\ref{sec:applications}).

\begin{figure}[!t]
    \centering\small
    \includegraphics[width=0.9\linewidth]{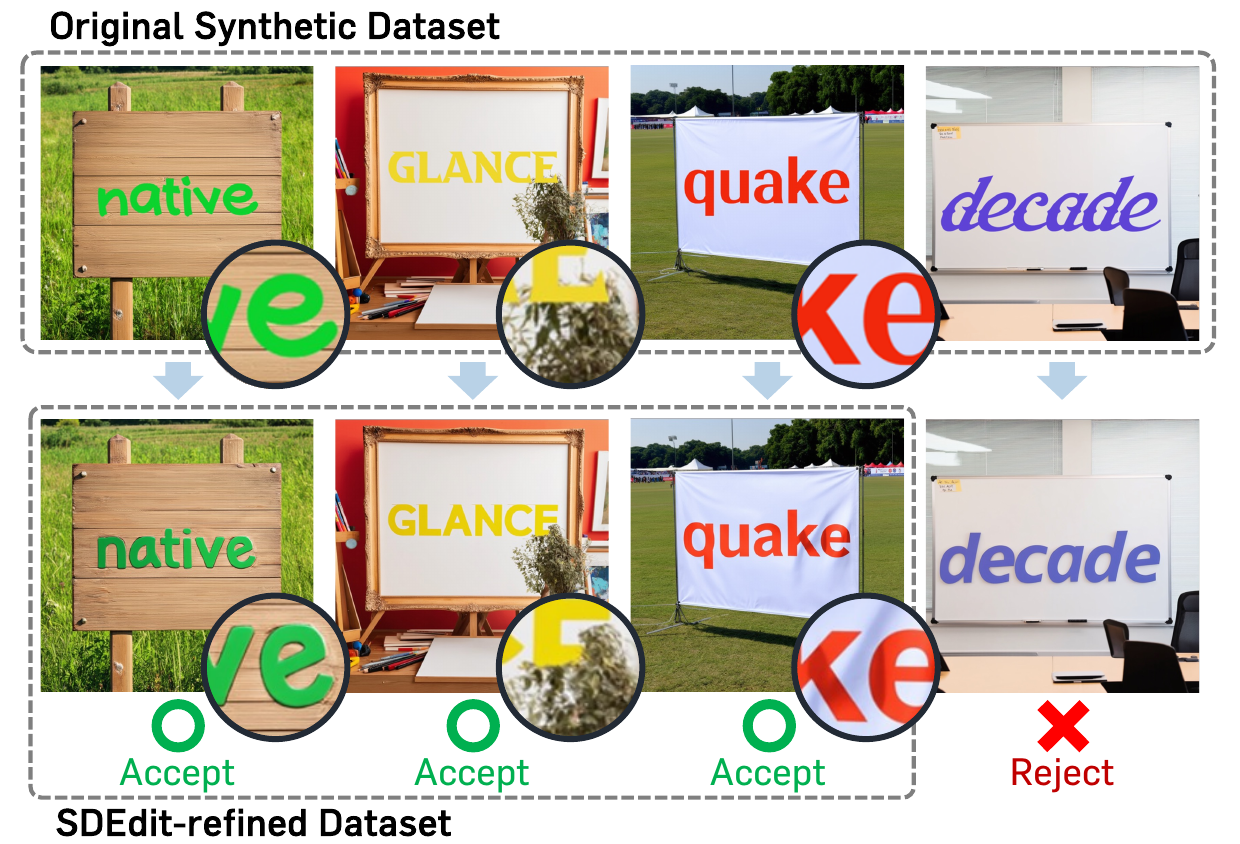}
    \vspace{-2mm}
    \caption{\textbf{Visual refinement using SDEdit.}
        Our original synthetic images (top row) are refined using SDEdit (bottom row).
        Low-quality images are filtered out based on font similarity.
    }
    \label{fig:sdedit}
    \vspace{-3mm}
\end{figure}
\begin{figure*}[t]
    \centering\small
    \includegraphics[width=\textwidth]{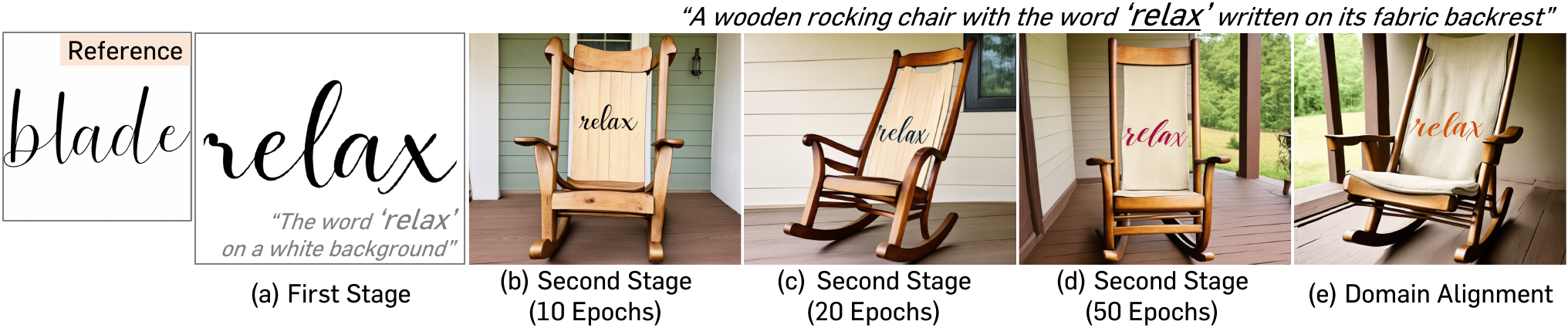}
    \captionsetup{justification=justified, singlelinecheck=false}
    \vspace{-6mm}
    \caption{\textbf{Font customization results over training.}
    Font attributes learned in the first stage remain consistent throughout training.
    }
    \vspace{-3mm}
    \label{fig:intermediates}
\end{figure*}

\subsection{\methodname}
\label{method:two_stage}

With our paired datasets designed for curriculum learning in font customization, this section details each stage of the two-stage training scheme and its distinct objectives.

\vspace{0.05in}
\noindent\textbf{First Stage: Extracting Font Attributes from Glyphs.}
In the first stage, \methodname focuses exclusively on the glyph itself, learning nuanced font styles shared between the reference and target images.
To achieve this, \methodname is trained using $(\mathbf{x}_\mathbf{t}, \mathbf{x}'_\mathbf{t})\in\mathcal{D}_\mathbf{t}$, where it generates a \textit{text-only} $\mathbf{x}'_\mathbf{t}$ from a reference glyph $\mathbf{x}_\mathbf{t}$ with different text content, both sharing a simple white background (see Figure~\ref{fig:dataset}).
During this stage, only the Resampler module and image-image attention layers are trained, while all other network parameters remain frozen. 
This ensures that \methodname effectively learns to extract font attributes from the image embeddings of the font reference, without modifying unrelated components.
This process is illustrated in Figure~\ref{fig:overview}(a).

\vspace{0.05in}
\noindent\textbf{Second Stage: Integrating Extracted Styles into Realistic Contexts.}
After the first stage, \methodname can generate new glyphs but only on simple white backgrounds.
To enable realistic visual text customization across diverse contexts, it is further trained with $(\mathbf{x}_\mathbf{t}, \mathbf{x}_\mathbf{st}) \in \mathcal{D}_\mathbf{st}$: \ie it generates a \textit{scene-text} image $\mathbf{x}_\mathbf{st}$ with varied backgrounds from a reference glyph $\mathbf{x}_\mathbf{t}$, which also differs in text content (see Figure~\ref{fig:dataset}).
At this stage, the Resampler module is frozen, and only the image-image attention layers are trained, as illustrated in Figure~\ref{fig:overview}(b).
This approach ensures realistic image generation while preserving the font attributes extracted by the Resampler, allowing seamless integration of visual text into diverse scenes.

\subsection{Additional Domain Alignment}
\label{method:domain_alignment}

While our two-stage training scheme effectively adapts font styles to image contexts (see Table~\ref{tab:ablation_components}), we introduce an additional domain alignment strategy to further enhance the realism of the synthetic dataset by incorporating real-world effects.
These include shadowing effects on glyphs (\eg \textit{``native''} in Figure~\ref{fig:sdedit}) and curvature distortions from complex textures (\eg \textit{``quake''} in Figure~\ref{fig:sdedit}).

We achieve this by employing SDEdit~\citep{meng2021sdedit}: adding noise to scene-text images, and using a T2I model to denoise glyph regions.
We then apply our proposed font accuracy metrics (see Section~\ref{sec:experiments}) to filter out excessively distorted outputs (see Figure~\ref{fig:sdedit} for examples).
Additionally, we generate high-quality samples using 50 expert models, each fine-tuned on 3–5 scene-text images, specializing in a single font style (see Appendix~\ref{suppl:datasets} for details and examples).

Finally, we fine-tune our model on these refined datasets while keeping the Resampler frozen, as in the second stage. This ensures that the model enhances realism while maintaining its ability to accurately extract font attributes.
\begin{figure*}[t]
    \centering\small
    \includegraphics[width =0.98\textwidth]{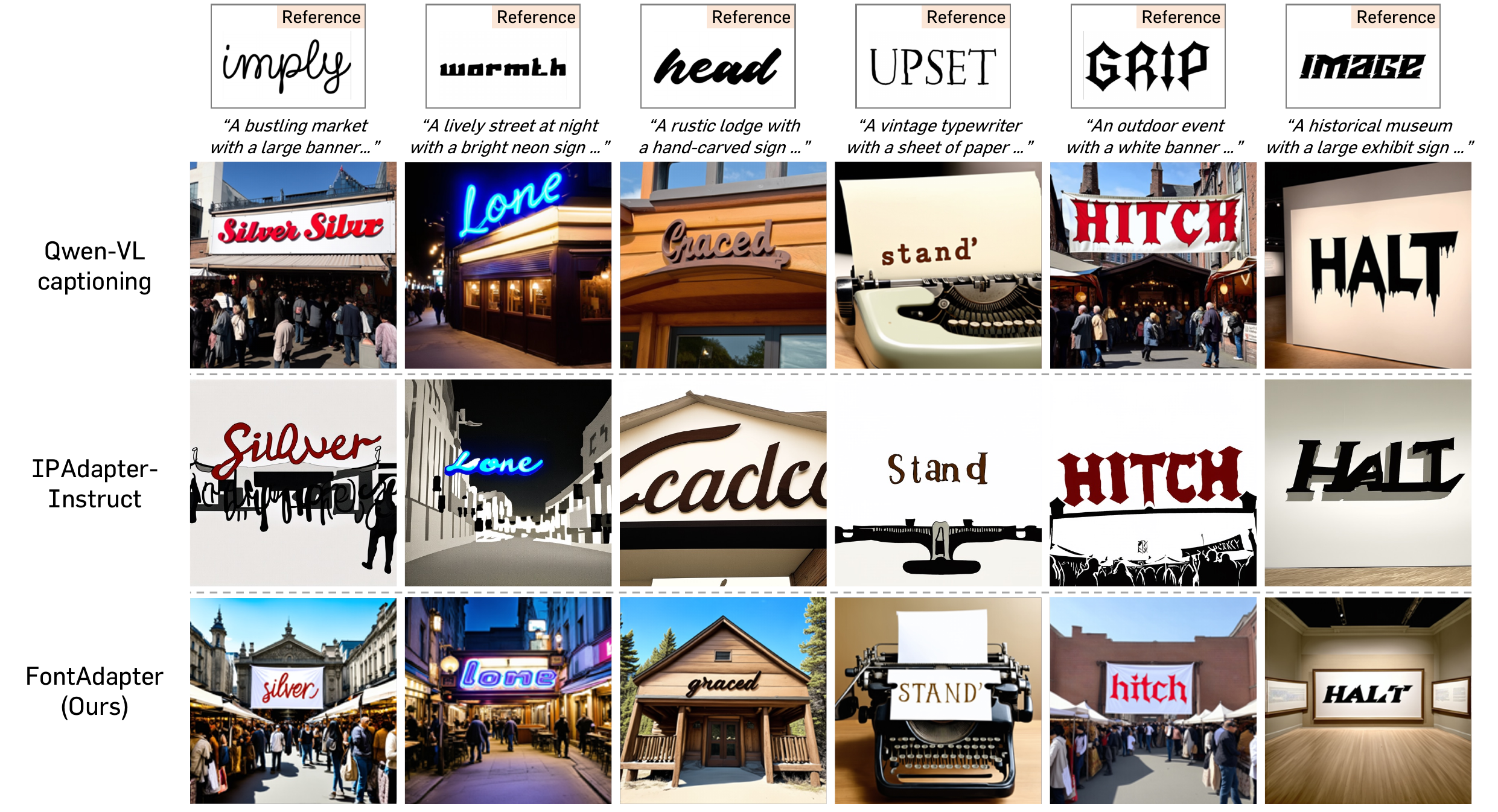}
    \vspace{-3mm}
    \caption{\textbf{Qualitative comparison of font customization.}
    Each sample is generated with a single font reference, without fine-tuning.
    \methodname demonstrates robust generalizability to diverse unique fonts while naturally integrating visual text into complex scenes.
    For example, in the first column, \methodname successfully generates visual text written in desired font style, well integrated with the realistic market scene.
    However, IPAdapter-Instruct struggles to adapt font attributes locally, resulting in a more graphic-like image, while Qwen-VL captioning struggles to adapt font style precisely, only resembling the desired font’s roundness.
    }
    \label{fig:qualitative}
    \vspace{-1mm}
\end{figure*}
\begin{table*}[t!]
\centering\small
\caption{\textbf{Quantitative comparison.}
    All samples are generated with 30 denoising steps.
    \methodname outperforms all baselines across font similarity metrics, effectively capturing and adapting font styles into new visual texts, while preserving the base model’s priors in text accuracy and prompt alignment, even with font conditioning.
    Notably, \methodname achieves higher text accuracy than its base model, indicating that our training method enables a stronger focus on generating precise visual text regions.
    \textbf{Bold} indicates the best result.
}
\label{tab:quantitative}

\vspace{-0.7em}

\resizebox{0.95\linewidth}{!}
{
\begin{tabular}{l c c c c c c c c}

    \toprule
    \multirow{2}{*}{Method} & \multicolumn{4}{c}{Font Similarity} & \multicolumn{2}{c}{Text Accuracy} & \multicolumn{2}{c}{Prompt Alignment} \\
    \cmidrule(lr){2-5} \cmidrule(lr){6-7} \cmidrule(lr){8-9}
    & {Max-IoU$\uparrow$} & {HOG-sim.$\uparrow$} & {MS-SSIM$\uparrow$} & {LPIPS$\downarrow$} & {WordAcc$\uparrow$} & {NED$\uparrow$} & {CLIP score$\uparrow$} & {SigLIP score$\uparrow$} \\
    \midrule
    SD3~\citep{esser2024scaling} & -      & -      & -      & -      & 0.3963 & 0.6472 & 31.50 & 0.9805 \\
    \midrule
    Qwen-VL captioning~\citep{bai2023qwen} & 0.3118 & 0.4451 & 0.3505 & 0.5915 & 0.2470 & 0.5031 & 30.72 & 0.9392 \\
    IPAdapter-Instruct~\citep{rowles2024ipadapter} & 0.3105 & 0.4481 & 0.6074 & 0.3321 & 0.0788 & 0.3686 & 25.86 & 0.6468 \\

    \rowcolor{Magenta!15}
    \textbf{\methodname (ours)} & \textbf{0.4293} & \textbf{0.6138} & \textbf{0.6899} & \textbf{0.2482} & \textbf{0.5303} & \textbf{0.7904} & \textbf{31.55} & \textbf{0.9816} \\
    \bottomrule

\end{tabular}
}
\vspace{-0.08in}
\end{table*}
\begin{table*}[t!]

\caption{\textbf{Component-wise ablations.}
    All training runs continue until performance saturation.
    Conducting the first stage beforehand boosts the model’s font and text accuracy ($2^\text{nd}$, $3^\text{rd}$ $\rightarrow$ $5^\text{th}$ row), while the domain alignment further expands performance boundaries ($5^\text{th}$ $\rightarrow$ $6^\text{th}$ row), enhancing prompt alignment as well as font and text accuracy.
    \textbf{Bold} and \underline{underline} indicates best and runner-up results, respectively. \\
    \dag Each sign denotes whether to freeze the Resampler on each stage. \textbf{(TO/ST: Text-Only/Scene-Text dataset, DA: Domain Alignment)}
}
\label{tab:ablation_components}

\vspace{-0.6em}

\resizebox{\linewidth}{!}
{
\begin{tabular}{c c c c c c c c c c}
    \toprule
    \raisebox{-.3\height}{Training} & \raisebox{-.3\height}{\dag Freeze} & \multicolumn{4}{c}{Font Similarity} & \multicolumn{2}{c}{Text Accuracy} & \multicolumn{2}{c}{Prompt Alignment} \\
    \cmidrule(lr){3-6} \cmidrule(lr){7-8} \cmidrule(lr){9-10}
    \raisebox{.3\height}{Scheme} & \raisebox{.3\height}{Resampler} & {Max-IoU$\uparrow$} & {HOG-sim.$\uparrow$} & {MS-SSIM$\uparrow$} & {LPIPS$\downarrow$} & {WordAcc$\uparrow$} & {NED$\uparrow$} & {CLIP score$\uparrow$} & {SigLIP score$\uparrow$} \\
    \midrule

    {Single-stage \small{(TO+ST)}} & \textcolor{RoyalBlue}{\cmark} & 0.3970 & 0.5805 & 0.6816 & 0.2601 & 0.4411 & 0.7197 & 31.31 & \textbf{0.9831} \\
    {Single-stage \small{(TO+ST)}} & \textcolor{magenta}{\xmark} & 0.4016 & 0.5950 & 0.6860 & 0.2543 & 0.3951 & 0.7066 & 31.29 & 0.9789 \\
    {Single-stage \small{(ST)}} & \textcolor{magenta}{\xmark} & 0.4036 & 0.5868 & 0.6796 & 0.2600 & 0.4554 & 0.7445 & \underline{31.42} & \underline{0.9818} \\
    \midrule

    {Two-stage} & \textcolor{magenta}{\xmark} / \textcolor{magenta}{\xmark} & 0.4180 & 0.6055 & 0.6861 & 0.2533 & 0.4384 & 0.7297 & 31.34 & 0.9776 \\
    {Two-stage} & \textcolor{magenta}{\xmark} / \textcolor{RoyalBlue}{\cmark} & \underline{0.4236} & \underline{0.6091} & \underline{0.6870} & \underline{0.2514} & \underline{0.4621} & \underline{0.7502} & 31.30 & 0.9747 \\

    \rowcolor{Magenta!15}
    {Two-stage + DA} & \textcolor{magenta}{\xmark} / \textcolor{RoyalBlue}{\cmark} & \textbf{0.4293} & \textbf{0.6138} & \textbf{0.6899} & \textbf{0.2482} & \textbf{0.5303} & \textbf{0.7904} & \textbf{31.55} & 0.9816 \\
    \bottomrule

\end{tabular}
}

\vspace{0.8em}

\centering\small
\caption{\textbf{Ablations on reference and target text pairing.}
    All the models are two-stage trained on the same datasets, with variations in the reference-target text pairs:
    the reference text is either same, different, or a mix (same 1 : different 3) relative to the target.
    The model performs best when trained with different text, underscoring the benefit of using synthetic datasets.
    \textbf{Bold} indicates the best result.
}
\label{tab:ablation_text_pair}

\vspace{-0.7em}

\resizebox{.9\linewidth}{!}
{
\begin{tabular}{c c c c c c c c c}
    \toprule
    \raisebox{-.3\height}{Text} & \multicolumn{4}{c}{Font Similarity} & \multicolumn{2}{c}{Text Accuracy} & \multicolumn{2}{c}{Prompt Alignment} \\
    \cmidrule(lr){2-5} \cmidrule(lr){6-7} \cmidrule(lr){8-9}
    \raisebox{.3\height}{Pairing} & {Max-IoU$\uparrow$} & {HOG-sim.$\uparrow$} & {MS-SSIM$\uparrow$} & {LPIPS$\downarrow$} & {WordAcc$\uparrow$} & {NED$\uparrow$} & {CLIP score$\uparrow$} & {SigLIP score$\uparrow$} \\
    \midrule

    Same      & 0.4071 & 0.5991 & 0.6848 & 0.2539 & 0.3177 & 0.6860 & 31.16 & 0.9744 \\
    Mixed     & 0.4167 & 0.6071 & \textbf{0.6870} & \textbf{0.2507} & 0.4065 & 0.7197 & 31.27 & 0.9719 \\
    \rowcolor{Magenta!15}
    Different & \textbf{0.4236} & \textbf{0.6091} & \textbf{0.6870} & 0.2514 & \textbf{0.4621} & \textbf{0.7502} & \textbf{31.30} & \textbf{0.9747} \\
    \bottomrule

\end{tabular}
}
\vspace{-0.5em}
\end{table*}

\section{Experiments}
\label{sec:experiments}

In this section, we validate the effectiveness of \methodname.
The results show that \methodname significantly outperforms the baseline, both qualitatively and quantitatively (Section~\ref{sec:main_results}).
Furthermore, we confirm the value of the proposed synthetic dataset and two-stage training scheme through ablation studies (Section~\ref{sec:ablations}).
We also show the potential for broader applications of \methodname, such as font blending and visual text editing (Section~\ref{sec:applications}).

\vspace{0.05in}
\noindent\textbf{Evaluation Metrics.}
We evaluate the model's performance based on three key aspects and their corresponding metrics. See Appendix~\ref{suppl:eval_metrics} for further details and visualizations.
\begin{itemize}
    \item \textit{Font accuracy} measures how well the model reproduces an unseen font style from a reference image. To assess this, we compute Max-IoU, HOG-similarity, MS-SSIM, and LPIPS after segmenting the generated visual text and aligning it with the ground truth glyph.
    \item \textit{Text accuracy} measures how well the model preserves the original text while adapting to unique font styles. To assess this, we compute Word Accuracy and NED \citep{tuo2023anytext, yang2023glyphcontrol}.
    \item \textit{Prompt alignment} evaluates how well the generated visual text aligns with the given text prompt. We use CLIP score and SigLIP score as metrics for this assessment.
\end{itemize}
For evaluation, we use a total of 9,000 prompts.
Specifically, we first collect 300 additional online fonts and compile a new dictionary of 1,000 English words to avoid overlap with the training set.
We then use GPT-4o~\citep{OpenAI2024ChatGPT4o} to generate 300 prompts for each image context complexity level---simple, moderate, and complex---resulting in a total of 900 prompts (see examples in Appendix~\ref{suppl:eval_prompts}).
For each evaluation font, we randomly sample 10 prompts, generating 9,000 samples in total for evaluation.

\vspace{0.05in}
\noindent\textbf{Implementation Details.}
Our model is initialized from the IPAdapter-Instruct~\citep{rowles2024ipadapter} model pretrained on SD3~\citep{esser2024scaling}, which is capable of visual text rendering.
Hence, we follow implementation details from IPAdapter-Instruct. 
Our training proceeds in an end-to-end manner, with 50 epochs for each of two-stage and an additional 10 epochs for the domain alignment.
We use the AdamW optimizer with a learning rate of $7.5\times10^{-8}$ for the two-stage training and $1\times10^{-8}$ for the domain alignment on a batch size of 48.

\vspace{0.05in}
\noindent\textbf{Baselines.}
We construct a baseline as the most viable alternative by using a vision-language model with T2I priors.
Specifically, we generate descriptive captions for each font style based on a reference image.
Here, we use Qwen-VL~\citep{bai2023qwen} to caption font styles, as it shows robust performance in reading visual text (see examples in Appendix~\ref{suppl:captioning}).
We then integrate these descriptions into the evaluation prompts for SD3 to generate samples, referred to as \textit{Qwen-VL captioning}.
We also consider IPAdapter-Instruct~\citep{rowles2024ipadapter} as another baseline to evaluate the generalizability of a tuning-free customization method.
We use a pretrained Hugging Face model that has not been further trained on font-specific datasets but still demonstrates strong adaptability to various image prompts, particularly when built on the text-generalizable T2I backbone, SD3.
Note that although the glyph image with a white background causes some image quality degradation, we find it to be the optimal choice for maximizing font style transfer while minimizing interference from other visual elements (details are in Appendix~\ref{suppl:ipadapter_instruct}).

\subsection{Experimental Results}
\label{sec:main_results}
\vspace{0.05in}
\noindent\textbf{Qualitative Results.}
As shown in Figure~\ref{fig:qualitative}, \methodname demonstrates remarkable performance in font customization across various font styles and image contexts (see full prompts in Appendix~\ref{suppl:more_examples_prompts}).
It effectively integrates visual text into realistic scenes, even for multiple words (see Figure~\ref{fig:teaser}), maintaining high consistency across them.
Meanwhile, we observe that font styles are largely lost when using captioning due to the limited expressiveness of textual descriptions and the T2I model’s weak ability to condition font styles through text prompts.
Additionally, while IPAdapter-Instruct tends to follow font styles when using our glyph images as references, it unintentionally introduces (monotonous) graphical artifacts into the outputs.

\vspace{0.05in}
\noindent\textbf{Quantitative Results.}
As shown in Table~\ref{tab:quantitative}, \methodname significantly outperforms the baseline models across all font accuracy metrics while maintaining text accuracy and prompt alignment of its base model.
Notably, our model demonstrates improved text accuracy even over its base model, SD3.
We hypothesize two contributing factors: first, training on glyph-specific datasets sharpens the model’s focus on rendering precise text, and second, the simplified visual contexts introduced through synthetic datasets provide more straightforward conditions for text rendering.
Additionally, our synthetic datasets lead our model to effectively adapt font attributes only while preserving other visual elements specified by prompts, resulting in higher \textit{CLIP score} and \textit{SigLIP score} compared to the baselines.

\subsection{Ablation Studies}
\label{sec:ablations}

\vspace{0.05in}
\noindent\textbf{Two-stage Training.}
To validate the necessity of our two-stage training scheme, we compare its performance against a na\"ive single-stage approach in Table~\ref{tab:ablation_components}.
Compared to training solely on the \textit{scene-text} dataset ($3^\text{rd}$ row), incorporating an initial stage on text-only images ($5^\text{th}$ row) noticeably enhances font similarity.
It also improves text accuracy, indicating that this stage helps the model properly adapt font styles while maintaining readability.
Moreover, training on both datasets in parallel ($2^\text{nd}$ row) even degrades font learning.
This confirms that the effectiveness of our two-stage training scheme does not merely stem from increased data exposure but from its carefully designed progression.

\begin{figure}[!t]
    \vspace{-2mm}
    \centering\small
    \includegraphics[width=0.95\linewidth]{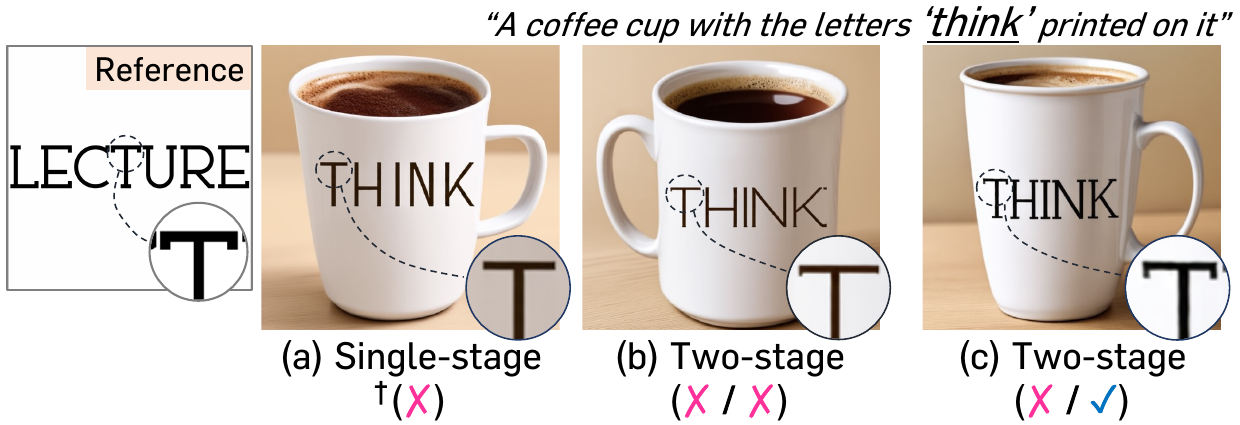}
    \vspace{-3mm}
    \caption{\textbf{Results from different training schemes.}
        Fine details of reference fonts stay preserved when the Resampler is frozen.
        \dag Each sign denotes whether to freeze the Resampler on each stage.
    }
    \vspace{-5mm}
    \label{fig:freeze_resampler}
\end{figure}

\vspace{0.05in}
\noindent\textbf{Resampler Freezing.}
We also examine the impact of freezing the Resampler by unfreezing it during the second stage ($4^\text{th}$ row).
Both font similarity and text accuracy decline when the Resampler is unfrozen, indicating that it loses previously learned font representations (also illustrated in Figure~\ref{fig:freeze_resampler}).
This is further supported by intermediate results in Figure~\ref{fig:intermediates}, where font details trained in the first stage remain consistent throughout subsequent training steps.
However, freezing the Resampler for the entire training process ($1^\text{st}$ row) severely degrades font similarity, highlighting the Resampler’s essential role in learning font attributes.

\vspace{0.05in}
\noindent\textbf{Domain Alignment.}
Interestingly, additional fine-tuning for domain alignment ($6^\text{th}$ row) not only enhances prompt alignment, as expected, but also improves font and text accuracy.
This suggests that the increased diversity in image context within our font-specific dataset enables the model to gain a deeper understanding of visual text rendering.

\vspace{0.05in}
\noindent\textbf{Reference and Target Text Pairing.}
We also design an ablation study to validate the effect of using the same or different text content in reference and target images, a key flexible option enabled by our synthetic datasets.
For the mixed setup, we use a 1:3 ratio of same-to-different pairs in the training dataset.
As shown in Table~\ref{tab:ablation_text_pair}, the model trained exclusively on different text content in the reference and target images achieves the best performance across nearly all metrics, highlighting the benefit of different text pairs in disentangling content-agnostic font attributes.

\begin{figure*}[!t]
    \centering\small
    \includegraphics[width=0.95\linewidth]{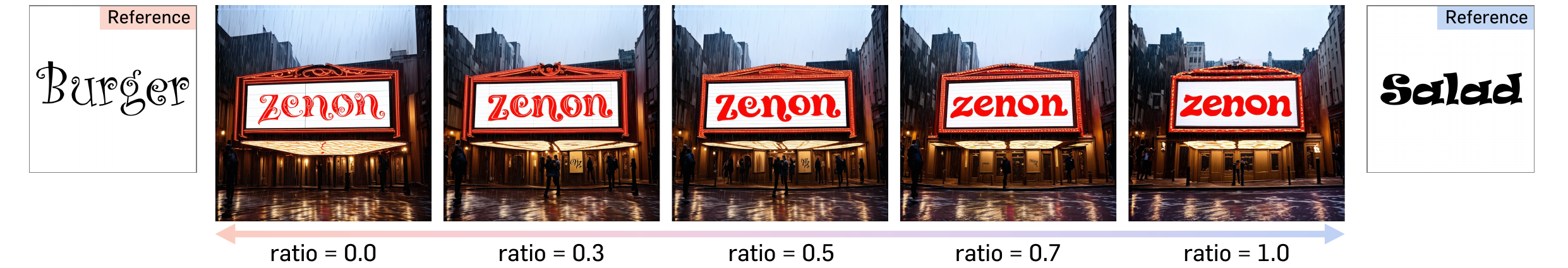}
    \vspace{-3mm}
    \caption{\textbf{Font blending.}
        Each sample between two font references is generated with interpolated image prompt tokens, where the ratio is indicated above each sample.
        Font features shift gradually according to the ratio, showing strong potential for designing new font styles.
    }
    \label{fig:app_blend}
    \vspace{-6mm}
\end{figure*}

\subsection{Applications}
\label{sec:applications}
We introduce additional noteworthy properties of \methodname beyond its font customization capabilities.

\begin{figure}[!t]
    \centering\small
    \includegraphics[width=0.85\linewidth]{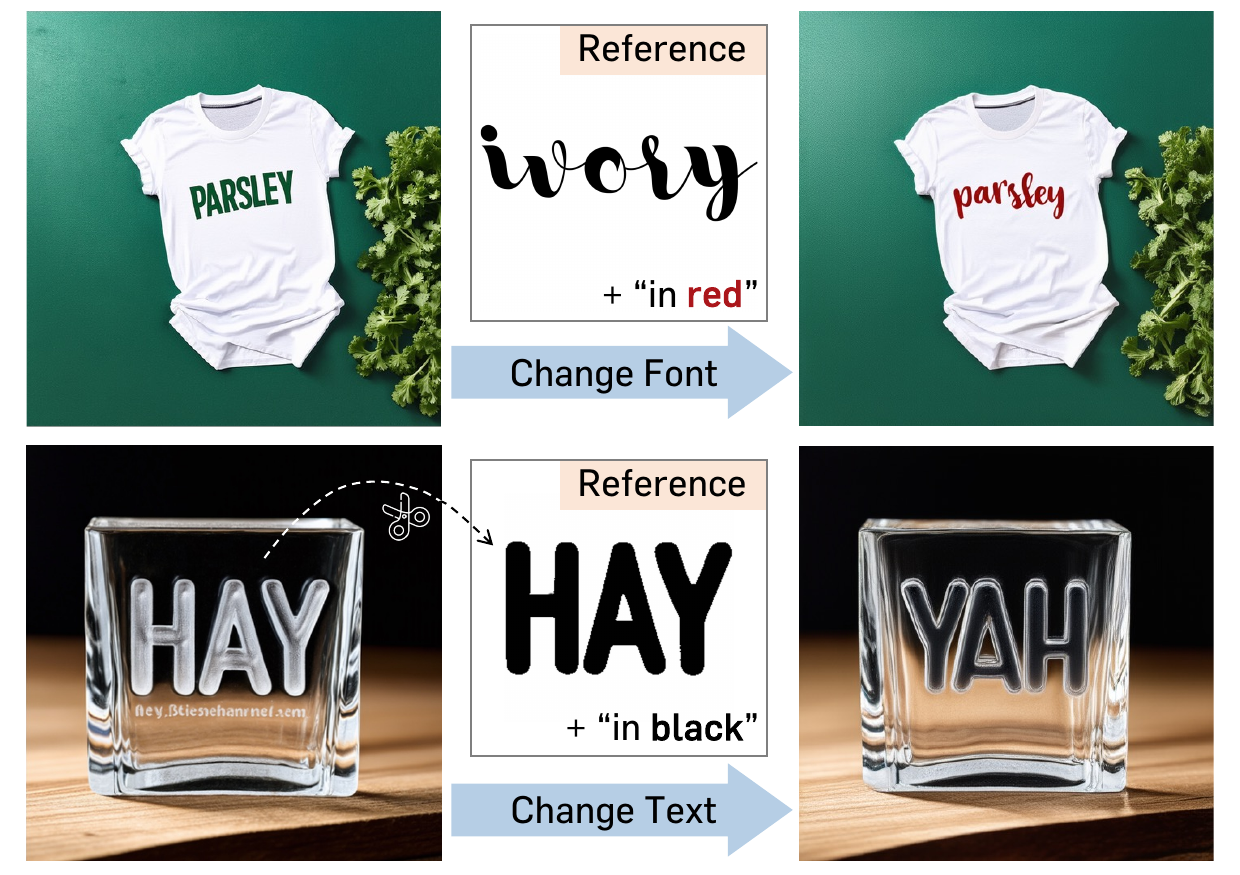}
    \vspace{-0.1in}
    \caption{\textbf{Visual text editing.}
        \methodname enables flexible control over text attributes for visual text editing using SDEdit.
        Control over font (top row) and text (bottom row) are presented.
    }
    \label{fig:app_edit}
    \vspace{-4mm}
\end{figure}

\vspace{0.05in}
\noindent\textbf{Visual Text Editing.}
\methodname enables flexible visual text editing using SDEdit~\citep{meng2021sdedit}.
As shown in Figure~\ref{fig:app_edit}, it allows modifying the text content, font style, or both, as well as adjusting its color through textual prompting.
Notably, when changing the text while preserving the font style, it can directly reference the original glyphs, with simple text segmentation.
This effectively addresses the limitation of T2I models in image editing, where font information is lost if the original glyph is masked for inpainting.

\vspace{0.05in}
\noindent\textbf{Font Blending.}
\methodname can also blend multiple font styles by interpolating their image prompt tokens (outputs from the Resampler), effectively preserving the distinct characteristics of each style.
As shown in Figure~\ref{fig:app_blend}, it smoothly transitions between two arbitrary font styles across a range of interpolation ratios, demonstrating its ability to generate novel hybrid styles that combine elements of both source fonts.
We expect this capability to offer users more flexible ways to customize font styles.

\begin{figure}[!t]
    \centering\small
    \includegraphics[width=0.85\linewidth]{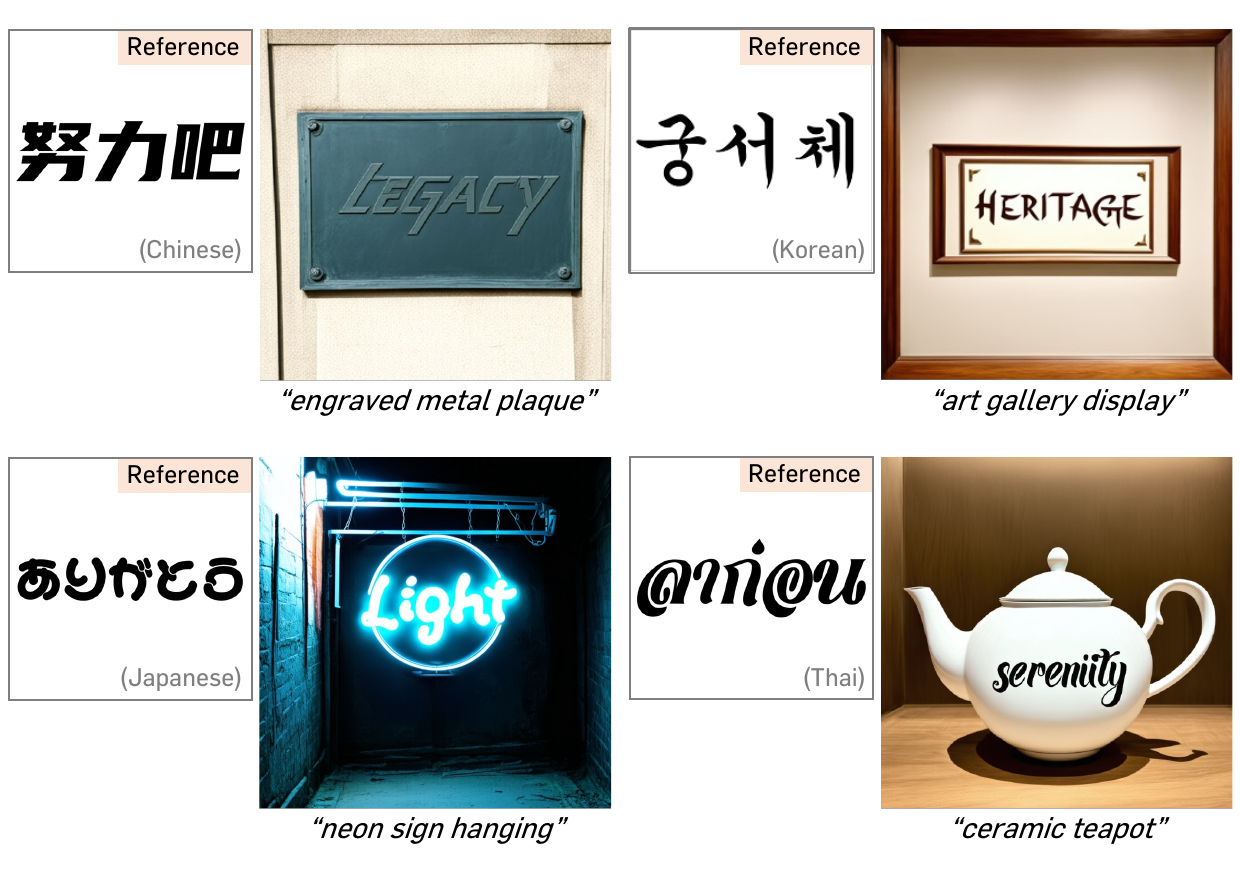}
    \vspace{-0.1in}
    \caption{\textbf{Cross-lingual font reference.}
        Although \methodname is not trained on other languages than English, it effectively captures font features from other language glyphs.
    }
    \label{fig:app_crossling}
    \vspace{-4mm}
\end{figure}

\vspace{0.05in}
\noindent\textbf{Cross-lingual Font Reference.} \methodname exhibits unique font customization capabilities, even in cross-lingual settings, as shown in Figure~\ref{fig:app_crossling}.  
This distinctive property arises from glyph-agnostic training of \methodname, where it is trained on font attributes using paired images with different glyphs. As a result, users can apply extracted font styles from one language to text in another, demonstrating its flexibility in multilingual font adaptation.

\section{Related Work}
\label{sec:related_work}

\vspace{0.05in}
\noindent\textbf{Font Customization.}
There has been growing interest in few-shot font generation~\citep{tang2022few, wang2023cf, pan2023few, fu2023neural, he2024diff, fu2024generate, yang2024fontdiffuser}, where the goal is to transfer a font style from a few reference glyphs to unseen characters.
With advances in visual text rendering through T2I models, several works have incorporated predefined font sets into fine-tuning to control font styles.
For example, DiffSTE~\citep{ji2023improving} fine-tunes an inpainting model with textual font descriptions for font-conditioned text editing, while Glyph-ByT5~\citep{liu2025glyph, liu2024glyph} uses special tokens and a glyph-specific text encoder for controlled design-text generation.
Although these models can render visual texts based on pretrained font styles, they lack generalizability to new (or user-specified) fonts.
Our approach, by contrast, focuses on instant font customization, allowing the instant generation of user-specified fonts using a reference image.

\vspace{0.05in}
\noindent\textbf{Dataset for Customization.}
To enable customization of specific subjects or styles on image generation, recent works have incorporated user-specified visual features from an additional input image into the generation process~\citep{wei2023elite, li2023blip, shi2024instantbooth, ye2023ip, he2024disenvisioner, rowles2024ipadapter}.
One line of research constructs datasets from video frames~\citep{chen2024anydoor, wang2024ms}, where the same object is captured from varied perspectives.
Another line of research leverages synthetic datasets generated by models, using methods such as constraining them to produce multiple images of a fixed identity~\citep{gal2024lcm, zeng2024jedi}, or providing detailed captions for a reference image using a multi-modal language model~\citep{he2024imagine}.

Besides, enabling image-driven control over font styles with such domain-specific datasets remains underexplored, where only non-visual font representations such as keywords~\citep{ji2023improving} or special tokens~\citep{liu2025glyph, liu2024glyph} have been employed to label specific fonts.
To address these limitations, we propose to construct a paired dataset consisting of visual text images and their corresponding font reference images.

\section{Conclusion}
\label{sec:conclusion}

This paper presents \methodname, a framework for efficient and flexible font customization in visual text generation.
Specifically, we propose a two-stage training scheme to capture detailed font attributes and diverse visual contexts, leveraging our synthetic font-specific datasets.
As a result, \methodname achieves high font accuracy and natural integration within realistic scenes.
We also show the versatility of \methodname for applications in image editing, font blending, and cross-language font transfer.
We hope this work contributes to advancing precise and intuitive control of text attributes in visual text generation.
\clearpage

{
    \small
    \bibliographystyle{ieeenat_fullname}
    \bibliography{main}

\begin{thebibliography}{43}
\providecommand{\natexlab}[1]{#1}
\providecommand{\url}[1]{\texttt{#1}}
\expandafter\ifx\csname urlstyle\endcsname\relax
  \providecommand{\doi}[1]{doi: #1}\else
  \providecommand{\doi}{doi: \begingroup \urlstyle{rm}\Url}\fi

\bibitem[Alayrac et~al.(2022)Alayrac, Donahue, Luc, Miech, Barr, Hasson, Lenc, Mensch, Millican, Reynolds, et~al.]{alayrac2022flamingo}
Jean-Baptiste Alayrac, Jeff Donahue, Pauline Luc, Antoine Miech, Iain Barr, Yana Hasson, Karel Lenc, Arthur Mensch, Katherine Millican, Malcolm Reynolds, et~al.
\newblock Flamingo: a visual language model for few-shot learning.
\newblock \emph{Advances in neural information processing systems}, 35:\penalty0 23716--23736, 2022.

\bibitem[Bai et~al.(2023)Bai, Bai, Yang, Wang, Tan, Wang, Lin, Zhou, and Zhou]{bai2023qwen}
Jinze Bai, Shuai Bai, Shusheng Yang, Shijie Wang, Sinan Tan, Peng Wang, Junyang Lin, Chang Zhou, and Jingren Zhou.
\newblock Qwen-vl: A versatile vision-language model for understanding, localization, text reading, and beyond.
\newblock \emph{arXiv preprint arXiv:2308.12966}, 1\penalty0 (2):\penalty0 3, 2023.

\bibitem[Betker et~al.(2023)Betker, Goh, Jing, Brooks, Wang, Li, Ouyang, Zhuang, Lee, Guo, et~al.]{betker2023improving}
James Betker, Gabriel Goh, Li Jing, Tim Brooks, Jianfeng Wang, Linjie Li, Long Ouyang, Juntang Zhuang, Joyce Lee, Yufei Guo, et~al.
\newblock Improving image generation with better captions.
\newblock \emph{Computer Science. https://cdn. openai. com/papers/dall-e-3. pdf}, 2\penalty0 (3):\penalty0 8, 2023.

\bibitem[{BlackForestLab}(2024)]{blackforestlab2024flux}
{BlackForestLab}.
\newblock Flux.1.
\newblock \url{https://blackforestlabs.ai/announcing-black-forest-labs/}, 2024.
\newblock Accessed: 2024-10-29.

\bibitem[Chen et~al.(2024)Chen, Huang, Liu, Shen, Zhao, and Zhao]{chen2024anydoor}
Xi Chen, Lianghua Huang, Yu Liu, Yujun Shen, Deli Zhao, and Hengshuang Zhao.
\newblock Anydoor: Zero-shot object-level image customization.
\newblock In \emph{Proceedings of the IEEE/CVF Conference on Computer Vision and Pattern Recognition}, pages 6593--6602, 2024.

\bibitem[Choi et~al.(2024)Choi, Kwak, Lee, Choi, and Shin]{choi2024improving}
Yisol Choi, Sangkyung Kwak, Kyungmin Lee, Hyungwon Choi, and Jinwoo Shin.
\newblock Improving diffusion models for virtual try-on.
\newblock \emph{arXiv preprint arXiv:2403.05139}, 2024.

\bibitem[DeepFloyd(2023)]{deepfloydif}
DeepFloyd.
\newblock Deepfloyd if.
\newblock \url{https://github.com/deep-floyd/IF}, 2023.

\bibitem[Esser et~al.(2024)Esser, Kulal, Blattmann, Entezari, M{\"u}ller, Saini, Levi, Lorenz, Sauer, Boesel, et~al.]{esser2024scaling}
Patrick Esser, Sumith Kulal, Andreas Blattmann, Rahim Entezari, Jonas M{\"u}ller, Harry Saini, Yam Levi, Dominik Lorenz, Axel Sauer, Frederic Boesel, et~al.
\newblock Scaling rectified flow transformers for high-resolution image synthesis.
\newblock In \emph{Forty-first International Conference on Machine Learning}, 2024.

\bibitem[Fu et~al.(2023)Fu, He, Wang, and Qiao]{fu2023neural}
Bin Fu, Junjun He, Jianjun Wang, and Yu Qiao.
\newblock Neural transformation fields for arbitrary-styled font generation.
\newblock In \emph{Proceedings of the IEEE/CVF conference on computer vision and pattern recognition}, pages 22438--22447, 2023.

\bibitem[Fu et~al.(2024)Fu, Yu, Liu, Wang, Wen, He, and Qiao]{fu2024generate}
Bin Fu, Fanghua Yu, Anran Liu, Zixuan Wang, Jie Wen, Junjun He, and Yu Qiao.
\newblock Generate like experts: Multi-stage font generation by incorporating font transfer process into diffusion models.
\newblock In \emph{Proceedings of the IEEE/CVF Conference on Computer Vision and Pattern Recognition}, pages 6892--6901, 2024.

\bibitem[Gal et~al.(2024)Gal, Lichter, Richardson, Patashnik, Bermano, Chechik, and Cohen-Or]{gal2024lcm}
Rinon Gal, Or Lichter, Elad Richardson, Or Patashnik, Amit~H Bermano, Gal Chechik, and Daniel Cohen-Or.
\newblock Lcm-lookahead for encoder-based text-to-image personalization.
\newblock \emph{arXiv preprint arXiv:2404.03620}, 2024.

\bibitem[He et~al.(2024{\natexlab{a}})He, Chen, Wang, Liu, Du, Tao, and Yu]{he2024diff}
Haibin He, Xinyuan Chen, Chaoyue Wang, Juhua Liu, Bo Du, Dacheng Tao, and Qiao Yu.
\newblock Diff-font: Diffusion model for robust one-shot font generation.
\newblock \emph{International Journal of Computer Vision}, pages 1--15, 2024{\natexlab{a}}.

\bibitem[He et~al.(2024{\natexlab{b}})He, Haodong, Shen, Yingjie, Qiu, Chen, et~al.]{he2024disenvisioner}
Jing He, LI Haodong, Guibao Shen, CAI Yingjie, Weichao Qiu, Ying-Cong Chen, et~al.
\newblock Disenvisioner: Disentangled and enriched visual prompt for customized image generation.
\newblock In \emph{The Thirteenth International Conference on Learning Representations}, 2024{\natexlab{b}}.

\bibitem[He et~al.(2024{\natexlab{c}})He, Sun, Juefei-Xu, Ma, Ramchandani, Cheung, Shah, Kalia, Subramanyam, Zareian, et~al.]{he2024imagine}
Zecheng He, Bo Sun, Felix Juefei-Xu, Haoyu Ma, Ankit Ramchandani, Vincent Cheung, Siddharth Shah, Anmol Kalia, Harihar Subramanyam, Alireza Zareian, et~al.
\newblock Imagine yourself: Tuning-free personalized image generation.
\newblock \emph{arXiv preprint arXiv:2409.13346}, 2024{\natexlab{c}}.

\bibitem[Hu et~al.(2021)Hu, Shen, Wallis, Allen-Zhu, Li, Wang, Wang, and Chen]{hu2021lora}
Edward~J Hu, Yelong Shen, Phillip Wallis, Zeyuan Allen-Zhu, Yuanzhi Li, Shean Wang, Lu Wang, and Weizhu Chen.
\newblock Lora: Low-rank adaptation of large language models.
\newblock \emph{arXiv preprint arXiv:2106.09685}, 2021.

\bibitem[Ji et~al.(2023)Ji, Zhang, Wang, Hou, Zhang, Price, and Chang]{ji2023improving}
Jiabao Ji, Guanhua Zhang, Zhaowen Wang, Bairu Hou, Zhifei Zhang, Brian Price, and Shiyu Chang.
\newblock Improving diffusion models for scene text editing with dual encoders.
\newblock \emph{arXiv preprint arXiv:2304.05568}, 2023.

\bibitem[Li et~al.(2022)Li, Liu, Guo, Yin, Jiang, Du, Du, Zhu, Lai, Hu, et~al.]{li2022pp}
Chenxia Li, Weiwei Liu, Ruoyu Guo, Xiaoting Yin, Kaitao Jiang, Yongkun Du, Yuning Du, Lingfeng Zhu, Baohua Lai, Xiaoguang Hu, et~al.
\newblock Pp-ocrv3: More attempts for the improvement of ultra lightweight ocr system.
\newblock \emph{arXiv preprint arXiv:2206.03001}, 2022.

\bibitem[Li et~al.(2023)Li, Li, and Hoi]{li2023blip}
Dongxu Li, Junnan Li, and Steven Hoi.
\newblock Blip-diffusion: Pre-trained subject representation for controllable text-to-image generation and editing.
\newblock \emph{Advances in Neural Information Processing Systems}, 36:\penalty0 30146--30166, 2023.

\bibitem[Liu et~al.(2022)Liu, Gong, and Liu]{liu2022flow}
Xingchao Liu, Chengyue Gong, and Qiang Liu.
\newblock Flow straight and fast: Learning to generate and transfer data with rectified flow.
\newblock \emph{arXiv preprint arXiv:2209.03003}, 2022.

\bibitem[Liu et~al.(2024)Liu, Liang, Zhao, Chen, Liang, Wang, Li, and Yuan]{liu2024glyph}
Zeyu Liu, Weicong Liang, Yiming Zhao, Bohan Chen, Lin Liang, Lijuan Wang, Ji Li, and Yuhui Yuan.
\newblock Glyph-byt5-v2: a strong aesthetic baseline for accurate multilingual visual text rendering.
\newblock \emph{arXiv preprint arXiv:2406.10208}, 2024.

\bibitem[Liu et~al.(2025)Liu, Liang, Liang, Luo, Li, Huang, and Yuan]{liu2025glyph}
Zeyu Liu, Weicong Liang, Zhanhao Liang, Chong Luo, Ji Li, Gao Huang, and Yuhui Yuan.
\newblock Glyph-byt5: A customized text encoder for accurate visual text rendering.
\newblock In \emph{European Conference on Computer Vision}, pages 361--377. Springer, 2025.

\bibitem[Meng et~al.(2021)Meng, He, Song, Song, Wu, Zhu, and Ermon]{meng2021sdedit}
Chenlin Meng, Yutong He, Yang Song, Jiaming Song, Jiajun Wu, Jun-Yan Zhu, and Stefano Ermon.
\newblock Sdedit: Guided image synthesis and editing with stochastic differential equations.
\newblock \emph{arXiv preprint arXiv:2108.01073}, 2021.

\bibitem[OpenAI(2023)]{dalle3}
OpenAI.
\newblock Dall-e 3: Text-to-image generation model.
\newblock \url{https://openai.com/dall-e-3}, 2023.

\bibitem[OpenAI(2024)]{OpenAI2024ChatGPT4o}
OpenAI.
\newblock Gpt-4o system card.
\newblock \url{https://openai.com/index/gpt-4o-system-card/}, 2024.

\bibitem[Pan et~al.(2023)Pan, Zhu, Zhou, Iwana, and Li]{pan2023few}
Wei Pan, Anna Zhu, Xinyu Zhou, Brian~Kenji Iwana, and Shilin Li.
\newblock Few shot font generation via transferring similarity guided global style and quantization local style.
\newblock In \emph{Proceedings of the IEEE/CVF International Conference on Computer Vision}, pages 19506--19516, 2023.

\bibitem[Radford et~al.(2021)Radford, Kim, Hallacy, Ramesh, Goh, Agarwal, Sastry, Askell, Mishkin, Clark, et~al.]{radford2021learning}
Alec Radford, Jong~Wook Kim, Chris Hallacy, Aditya Ramesh, Gabriel Goh, Sandhini Agarwal, Girish Sastry, Amanda Askell, Pamela Mishkin, Jack Clark, et~al.
\newblock Learning transferable visual models from natural language supervision.
\newblock In \emph{International conference on machine learning}, pages 8748--8763. PMLR, 2021.

\bibitem[Rombach et~al.(2022)Rombach, Blattmann, Lorenz, Esser, and Ommer]{rombach2022high}
Robin Rombach, Andreas Blattmann, Dominik Lorenz, Patrick Esser, and Bj{\"o}rn Ommer.
\newblock High-resolution image synthesis with latent diffusion models.
\newblock In \emph{Proceedings of the IEEE/CVF conference on computer vision and pattern recognition}, pages 10684--10695, 2022.

\bibitem[Rowles et~al.(2024)Rowles, Vainer, De~Nigris, Elizarov, Kutsy, and Donn{\'e}]{rowles2024ipadapter}
Ciara Rowles, Shimon Vainer, Dante De~Nigris, Slava Elizarov, Konstantin Kutsy, and Simon Donn{\'e}.
\newblock Ipadapter-instruct: Resolving ambiguity in image-based conditioning using instruct prompts.
\newblock \emph{arXiv preprint arXiv:2408.03209}, 2024.

\bibitem[Saharia et~al.(2022)Saharia, Chan, Saxena, Li, Whang, Denton, Ghasemipour, Gontijo~Lopes, Karagol~Ayan, Salimans, et~al.]{saharia2022photorealistic}
Chitwan Saharia, William Chan, Saurabh Saxena, Lala Li, Jay Whang, Emily~L Denton, Kamyar Ghasemipour, Raphael Gontijo~Lopes, Burcu Karagol~Ayan, Tim Salimans, et~al.
\newblock Photorealistic text-to-image diffusion models with deep language understanding.
\newblock \emph{Advances in neural information processing systems}, 35:\penalty0 36479--36494, 2022.

\bibitem[Shi et~al.(2024)Shi, Xiong, Lin, and Jung]{shi2024instantbooth}
Jing Shi, Wei Xiong, Zhe Lin, and Hyun~Joon Jung.
\newblock Instantbooth: Personalized text-to-image generation without test-time finetuning.
\newblock In \emph{Proceedings of the IEEE/CVF Conference on Computer Vision and Pattern Recognition}, pages 8543--8552, 2024.

\bibitem[Tang et~al.(2022)Tang, Cai, Liu, Hong, Gong, Fan, Han, Liu, Ding, and Wang]{tang2022few}
Licheng Tang, Yiyang Cai, Jiaming Liu, Zhibin Hong, Mingming Gong, Minhu Fan, Junyu Han, Jingtuo Liu, Errui Ding, and Jingdong Wang.
\newblock Few-shot font generation by learning fine-grained local styles.
\newblock In \emph{Proceedings of the IEEE/CVF conference on computer vision and pattern recognition}, pages 7895--7904, 2022.

\bibitem[Tuo et~al.(2023)Tuo, Xiang, He, Geng, and Xie]{tuo2023anytext}
Yuxiang Tuo, Wangmeng Xiang, Jun-Yan He, Yifeng Geng, and Xuansong Xie.
\newblock Anytext: Multilingual visual text generation and editing.
\newblock \emph{arXiv preprint arXiv:2311.03054}, 2023.

\bibitem[Van Den~Oord et~al.(2017)Van Den~Oord, Vinyals, et~al.]{van2017neural}
Aaron Van Den~Oord, Oriol Vinyals, et~al.
\newblock Neural discrete representation learning.
\newblock \emph{Advances in neural information processing systems}, 30, 2017.

\bibitem[Wang et~al.(2023)Wang, Zhou, Ge, Jiang, Bao, and Xu]{wang2023cf}
Chi Wang, Min Zhou, Tiezheng Ge, Yuning Jiang, Hujun Bao, and Weiwei Xu.
\newblock Cf-font: Content fusion for few-shot font generation.
\newblock In \emph{Proceedings of the IEEE/CVF Conference on Computer Vision and Pattern Recognition}, pages 1858--1867, 2023.

\bibitem[Wang et~al.(2020)Wang, Sun, Cheng, Jiang, Deng, Zhao, Liu, Mu, Tan, Wang, et~al.]{wang2020deep}
Jingdong Wang, Ke Sun, Tianheng Cheng, Borui Jiang, Chaorui Deng, Yang Zhao, Dong Liu, Yadong Mu, Mingkui Tan, Xinggang Wang, et~al.
\newblock Deep high-resolution representation learning for visual recognition.
\newblock \emph{IEEE transactions on pattern analysis and machine intelligence}, 43\penalty0 (10):\penalty0 3349--3364, 2020.

\bibitem[Wang et~al.(2024)Wang, Fu, Huang, He, and Jiang]{wang2024ms}
X Wang, Siming Fu, Qihan Huang, Wanggui He, and Hao Jiang.
\newblock Ms-diffusion: Multi-subject zero-shot image personalization with layout guidance.
\newblock \emph{arXiv preprint arXiv:2406.07209}, 2024.

\bibitem[Wei et~al.(2023)Wei, Zhang, Ji, Bai, Zhang, and Zuo]{wei2023elite}
Yuxiang Wei, Yabo Zhang, Zhilong Ji, Jinfeng Bai, Lei Zhang, and Wangmeng Zuo.
\newblock Elite: Encoding visual concepts into textual embeddings for customized text-to-image generation.
\newblock In \emph{Proceedings of the IEEE/CVF International Conference on Computer Vision}, pages 15943--15953, 2023.

\bibitem[Xu et~al.(2021)Xu, Zhang, Wang, Price, Wang, and Shi]{xu2021rethinking}
Xingqian Xu, Zhifei Zhang, Zhaowen Wang, Brian Price, Zhonghao Wang, and Humphrey Shi.
\newblock Rethinking text segmentation: A novel dataset and a text-specific refinement approach.
\newblock In \emph{Proceedings of the IEEE/CVF conference on computer vision and pattern recognition}, pages 12045--12055, 2021.

\bibitem[Yang et~al.(2023)Yang, Gui, Yuan, Liang, Ding, Hu, and Chen]{yang2023glyphcontrol}
Yukang Yang, Dongnan Gui, Yuhui Yuan, Weicong Liang, Haisong Ding, Han Hu, and Kai Chen.
\newblock Glyphcontrol: glyph conditional control for visual text generation.
\newblock \emph{Advances in Neural Information Processing Systems}, 36:\penalty0 44050--44066, 2023.

\bibitem[Yang et~al.(2024)Yang, Peng, Kong, Zhang, Yao, and Jin]{yang2024fontdiffuser}
Zhenhua Yang, Dezhi Peng, Yuxin Kong, Yuyi Zhang, Cong Yao, and Lianwen Jin.
\newblock Fontdiffuser: One-shot font generation via denoising diffusion with multi-scale content aggregation and style contrastive learning.
\newblock In \emph{Proceedings of the AAAI conference on artificial intelligence}, pages 6603--6611, 2024.

\bibitem[Ye et~al.(2023)Ye, Zhang, Liu, Han, and Yang]{ye2023ip}
Hu Ye, Jun Zhang, Sibo Liu, Xiao Han, and Wei Yang.
\newblock Ip-adapter: Text compatible image prompt adapter for text-to-image diffusion models.
\newblock \emph{arXiv preprint arXiv:2308.06721}, 2023.

\bibitem[Zeng et~al.(2024)Zeng, Patel, Wang, Huang, Wang, Liu, and Balaji]{zeng2024jedi}
Yu Zeng, Vishal~M Patel, Haochen Wang, Xun Huang, Ting-Chun Wang, Ming-Yu Liu, and Yogesh Balaji.
\newblock Jedi: Joint-image diffusion models for finetuning-free personalized text-to-image generation.
\newblock In \emph{Proceedings of the IEEE/CVF Conference on Computer Vision and Pattern Recognition}, pages 6786--6795, 2024.

\bibitem[Zhai et~al.(2023)Zhai, Mustafa, Kolesnikov, and Beyer]{zhai2023sigmoid}
Xiaohua Zhai, Basil Mustafa, Alexander Kolesnikov, and Lucas Beyer.
\newblock Sigmoid loss for language image pre-training.
\newblock In \emph{Proceedings of the IEEE/CVF International Conference on Computer Vision}, pages 11975--11986, 2023.

\end{thebibliography}
}

\clearpage
\appendix
\clearpage
\maketitlesupplementary

\section{Font-specific Datasets}
\label{suppl:datasets}

\begin{figure*}[t]
    \centering\small
    \includegraphics[width = \textwidth]{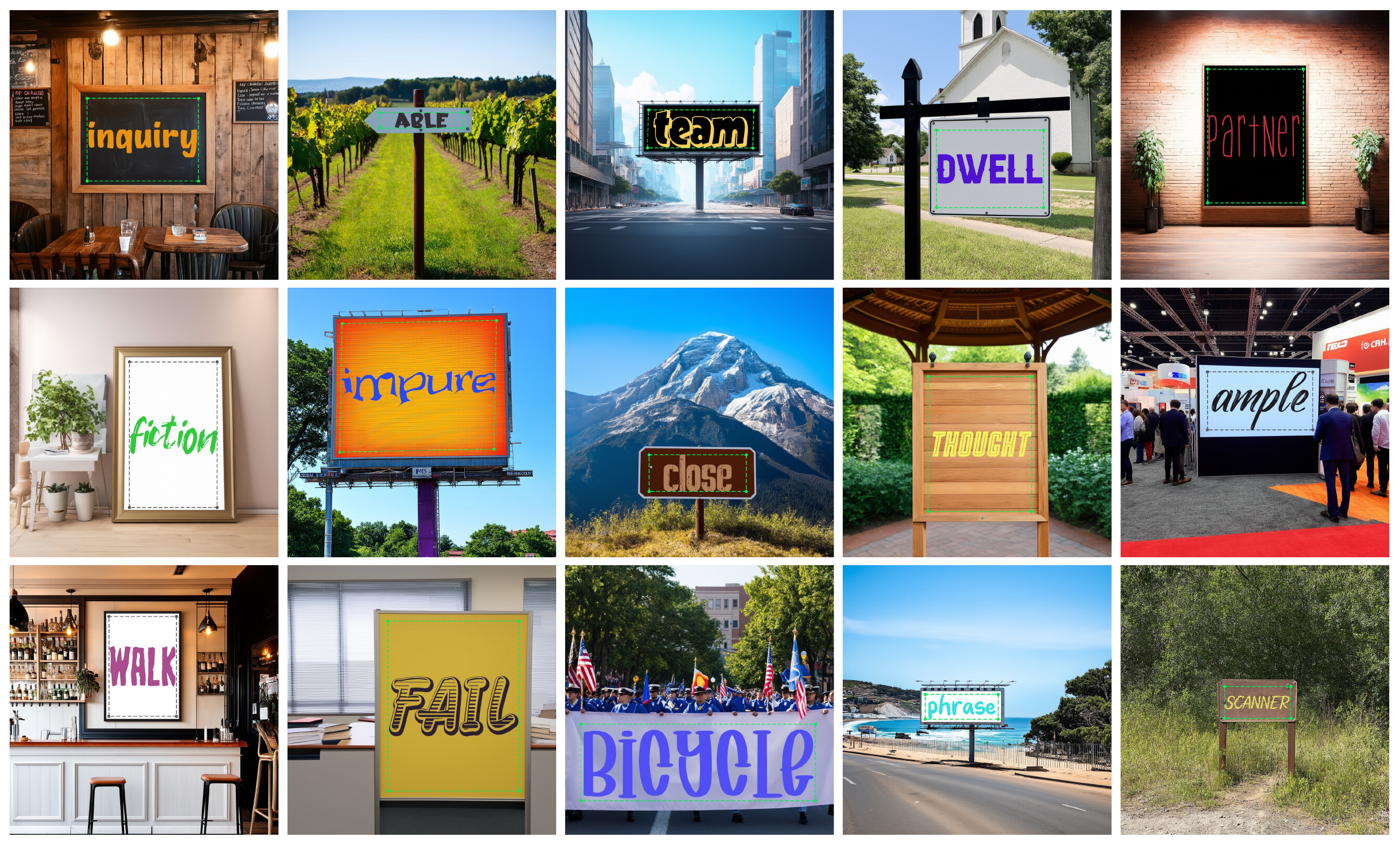}
    \caption{\textbf{Examples of scene-text images.}
    Each background image is generated using SD3 and manually labeled with corner coordinates of a text region, marked with dashed lines.
    The glyphs are then warped onto the bounding box, while preserving their aspect ratio.
    }
    \label{fig:suppl_scene_text}
\end{figure*}

\vspace{0.05in}
\noindent\textbf{Synthetic Datasets.}
To create the proposed \textit{synthetic dataset used for two-stage training}, we collect 1,500 online fonts and generate 500 realistic background images using SD3~\citep{esser2024scaling}.
We build a dictionary of 1,000 English words, each 4–7 characters long, and randomly sample 10 words per font.
Using this, we employ the Python Imaging Library to render black glyphs on a white background to generate a total of 15,000 \textit{text-only} images.
To generate \textit{scene-text} images, we randomly colorize these glyphs and overlay them onto backgrounds, as illustrated in Figure~\ref{fig:suppl_scene_text}.
The manually labeled bounding boxes for text placement are marked with dashed lines.
Additionally, detailed prompts used to generate the backgrounds are provided in Table~\ref{tab:suppl_bg_prompts}, along with the visual text descriptions added for training.
Each \textit{text-only} image serves as a font reference and is randomly paired with another \textit{text-only} or \textit{scene-text} image of the same font to form $\mathcal{D}_\mathbf{t}$ and $\mathcal{D}_\mathbf{st}$, respectively.

\begin{figure*}[t]
    \centering\small
    \includegraphics[width = \textwidth]{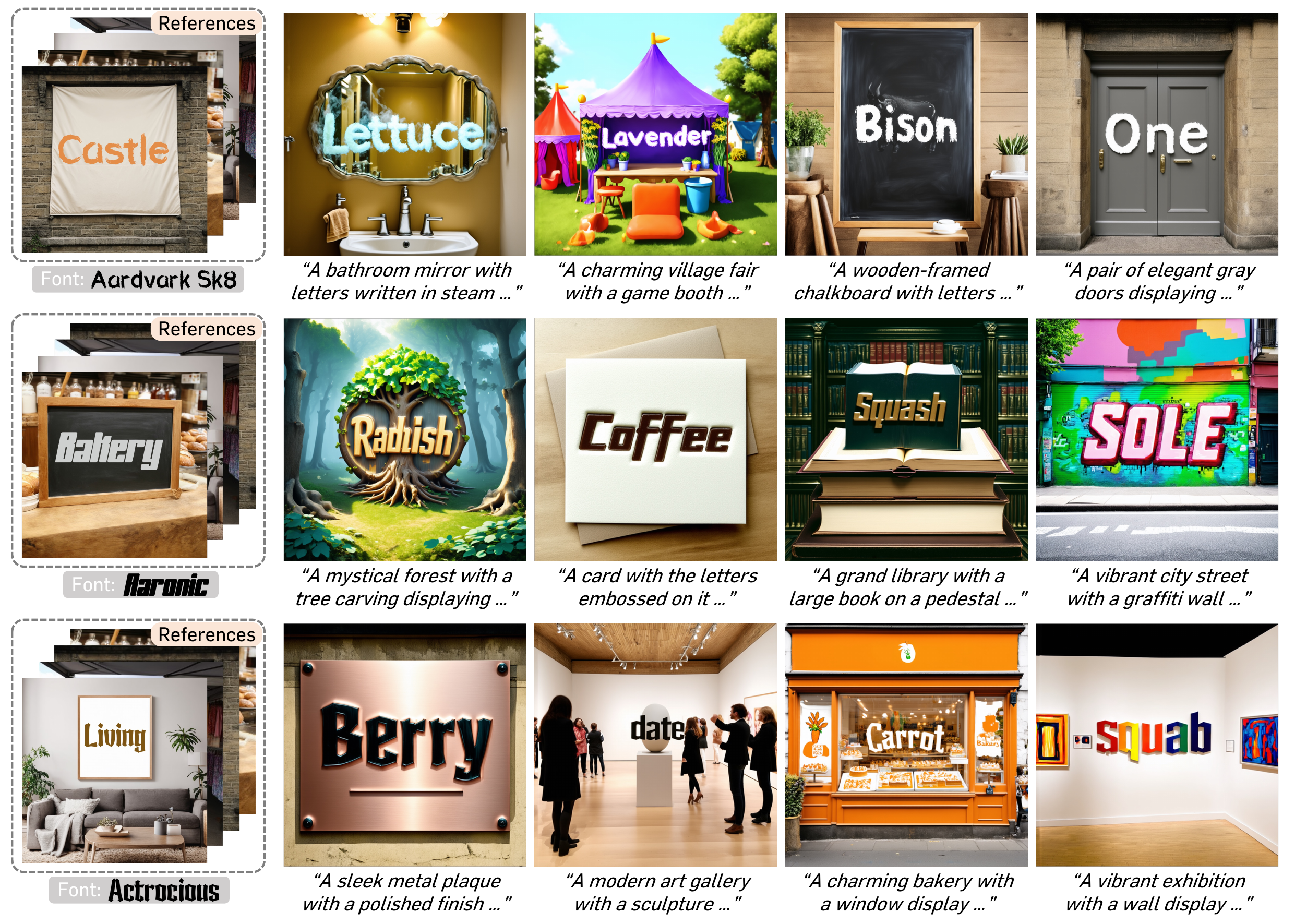}
    \vspace{-2.3em}
    \caption{\textbf{Examples of expert samples.}
    Each expert model is fine-tuned using 4 visual text images of a specific font (shown in the first column).
    The rest of the samples in each row are generated by the corresponding expert model, which demonstrate consistent font styles seamlessly integrated into diverse image contexts.
    These samples are used for additional fine-tuning for domain alignment.
    }
    \label{fig:suppl_expert_samples}
\end{figure*}

\vspace{0.05in}
\noindent\textbf{Additional Refined Datasets.}
For \textit{domain alignment} (Section~\ref{method:domain_alignment}), we refine the synthetic dataset using (1) SDEdit and (2) expert models.
(1) For SDEdit-based refinement, we select a subset of synthetic samples and refine them using strength values \{6, 7.5\}, which represent the ratio of added noise to the total denoising process.
To prevent undesirable alterations to font attributes, we filter the refined samples using the heuristic criterion ``{$\texttt{Max-IoU}>0.59 \land \texttt{HOG-sim.}>0.80$}'', which effectively excludes those with noticeable glyph deformation, resulting in 1,436 retained samples.
(2) For expert model training, we randomly select 50 fonts from the training dataset and fine-tune the models using LoRA~\citep{hu2021lora} with 4 \textit{scene-text} images per font.
Figure~\ref{fig:suppl_expert_samples} shows visual text images generated by expert models, demonstrating diverse image contexts while preserving font consistency.
We generate 10 samples per expert model, yielding a total of 500 samples.

\section{Evaluation Prompts}
\label{suppl:eval_prompts}

Examples of evaluation prompts are listed in Table~\ref{tab:suppl_eval_prompts}, for each complexity level of the image context.
Specifically, each \textsc{Simple} prompt describes only a single object with visual text written on it, \textsc{Moderate} prompts include descriptions of the background, and \textsc{Complex} prompts add additional context, such as another object beside it.
Additionally, each prompt specifies the color of the visual text to evaluate controllability over text colors as a separate attribute, ensuring it is disentangled from font styles.

\section{Evaluation Pipeline and Metrics}
\label{suppl:eval_metrics}

\begin{figure*}[t]
    \centering\small
    \includegraphics[width = 0.9\textwidth]{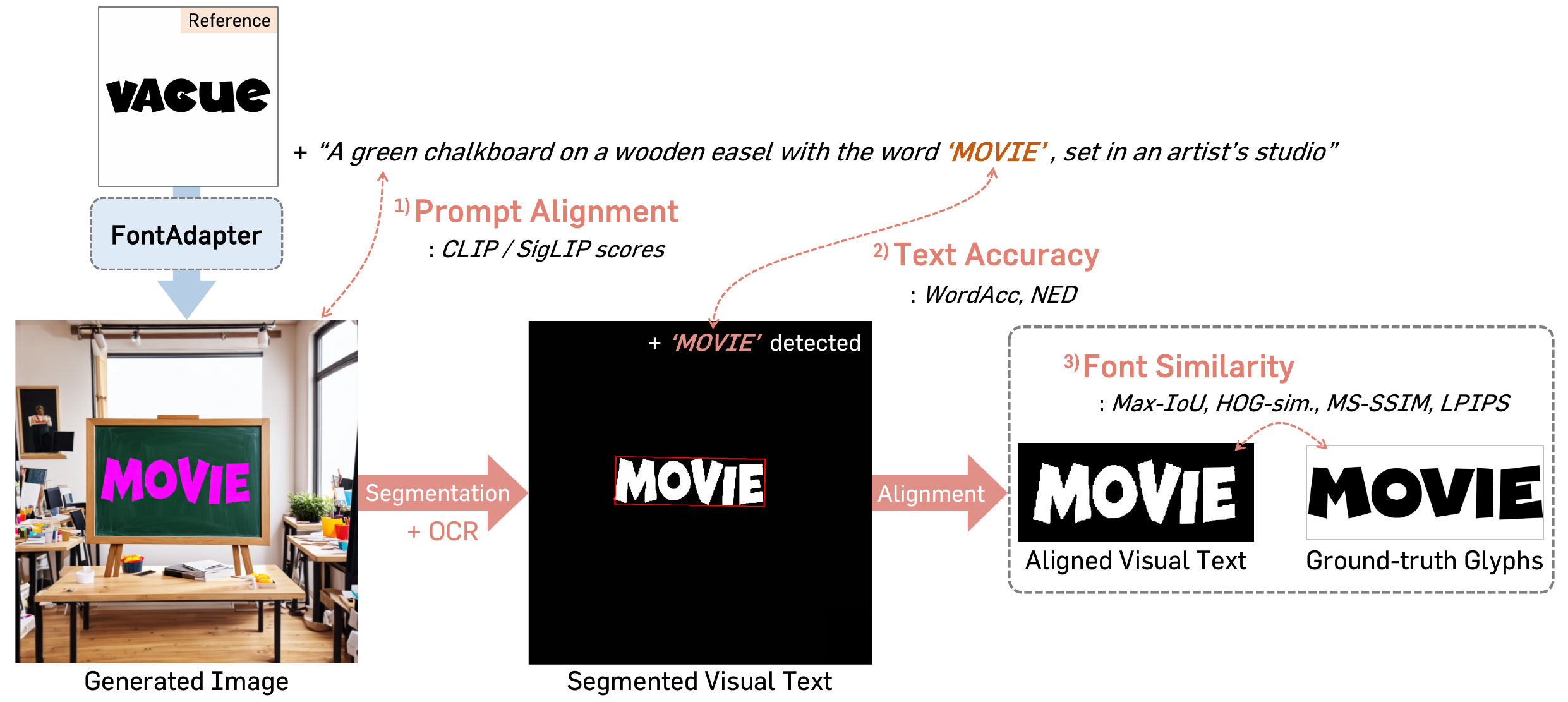}
    \vspace{-1em}
    \caption{\textbf{Visualization of evaluation pipeline.}
    First, $^{1)}$prompt alignment is assessed by measuring image-text similarity between the generated image and its input prompt.
    Next, $^{2)}$text accuracy is evaluated using an off-the-shelf OCR model (PP-OCRv3~\citep{li2022pp}).
    Lastly, for $^{3)}$font similarity, the visual text is segmented at the pixel level using a text segmentation model (TexRNet~\citep{xu2021rethinking}) and aligned with its ground-truth glyphs by maximizing IoU values.
    Once aligned, font similarity metrics are computed between the two aligned glyph images.
    }
    \label{fig:suppl_eval_pipeline}
\end{figure*}

\vspace{0.05in}
\noindent\textbf{Overall Evaluation Pipeline.}
We illustrate the evaluation pipeline in Figure~\ref{fig:suppl_eval_pipeline}.
First, prompt alignment is evaluated by measuring the image-text similarity between the generated image and its input prompt.
Next, text accuracy is assessed by recognizing the spelling of the visual text using PP-OCRv3~\citep{li2022pp}, an off-the-shelf OCR model.
Lastly, font similarity is measured by preprocessing the visual text: segmenting it from its background using TexRNet~\citep{xu2021rethinking} based on HRNet~\citep{wang2020deep}, and aligning it with the ground-truth glyphs by maximizing the Intersection over Union (IoU).
The alignment is performed via a sweeping search, allowing only translation and scaling to preserve the font style.
We infer this maximized IoU value as ``Max-IoU'', which effectively captures the spatial alignment between two glyphs.
Along with it, we also measure \textit{HOG-similarity}, a cosine distance between the Histogram of Oriented Gradients (HOG) to capture the boundary alignment.
For text accuracy, we measure Word Accuracy (\textit{WordAcc}), a word-level binary comparison, and Normalized Edit Distance (\textit{NED}), a character-level distance metric.
Both are case-agnostic metrics widely used in visual text generation literature~\citep{tuo2023anytext, yang2023glyphcontrol}, evaluated under varying strictness.
Finally, to assess how faithfully the model follows prompts, we use two well-known metrics for image-text alignment, \ie \textit{CLIP score}~\citep{radford2021learning} and \textit{SigLIP score}~\citep{zhai2023sigmoid}.

\begin{figure*}[t]
    \centering\small
    \includegraphics[width=1.0\textwidth]{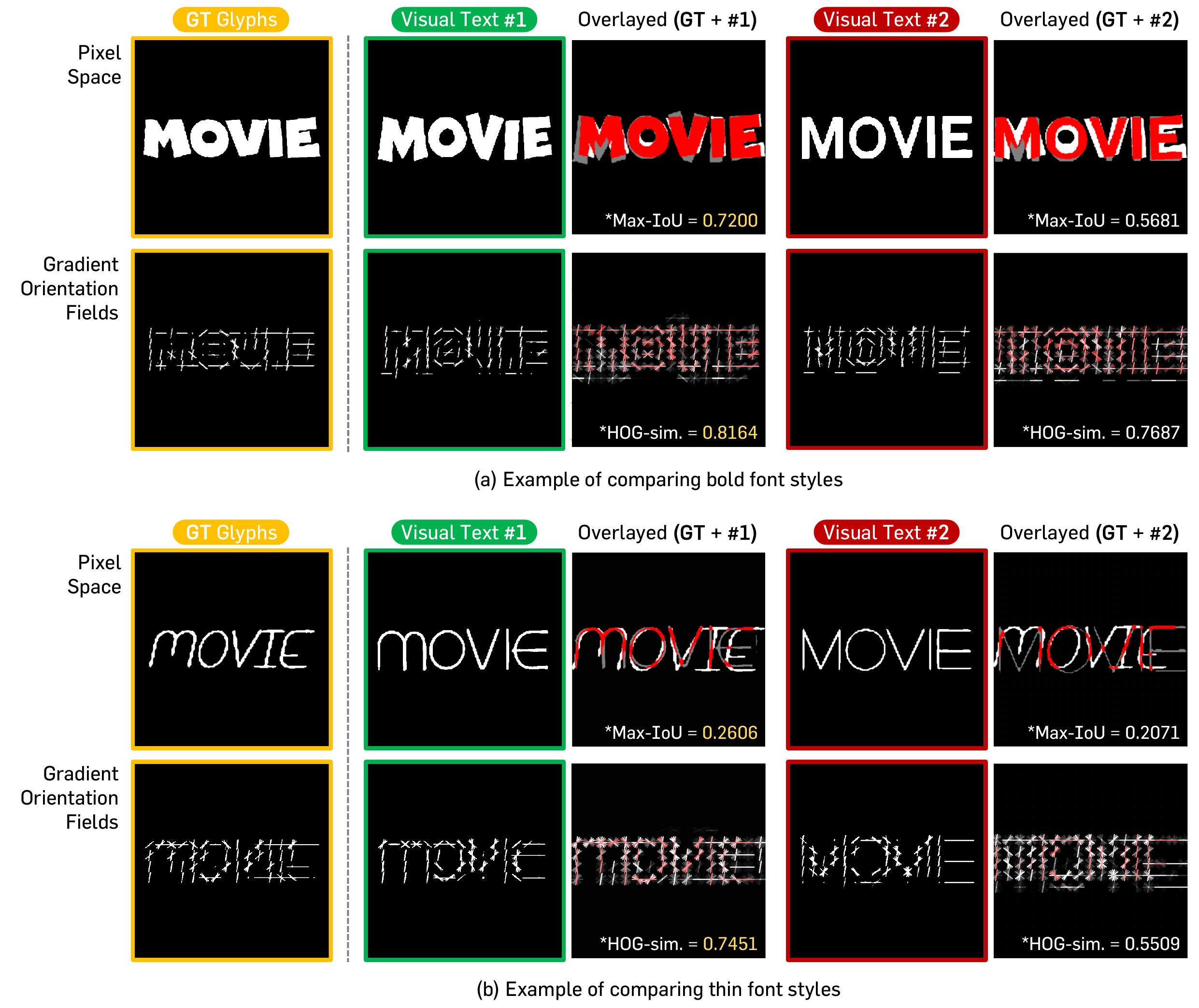}
    \caption{\textbf{Examples of measuring max-IoU and HOG-similarity.}
    Ground-truth glyphs are highlighted in \sethlcolor{Goldenrod!50}\hl{yellow}, with the more similar visual text (\#1) in \sethlcolor{ForestGreen!30}\hl{green} and the less similar one (\#2) in \sethlcolor{Maroon!30}\hl{red}.
    (a) and (b) show examples of bold and thin font styles, respectively.
    In both cases, the first row aligns each visual text with the ground-truth glyphs by maximizing IoU (overlapped regions highlighted in red), while the second row visualizes HOG features as gradient orientation fields (overlapping components highlighted in \textcolor{red}{\textbf{red}}).
    Both of max-IoU and HOG-similarity effectively measure font similarity, consistently assigning higher values to the visual text \#1.
    Notably, HOG-similarity remains robust in its absolute values, even for thin font styles as shown in (b).
    }
    \label{fig:suppl_eval_metrics}
\end{figure*}

\vspace{0.05in}
\noindent\textbf{Font Similarity Metrics.}
We visualize how the font similarity metrics, especially \textit{max-IoU} and \textit{HOG-similarity}, effectively compare two glyphs.
To demonstrate their robustness across glyph thicknesses, we present bold fonts in Figure~\ref{fig:suppl_eval_metrics}(a) and thin fonts in Figure~\ref{fig:suppl_eval_metrics}(b).
Relative to ground-truth glyphs (in \sethlcolor{Goldenrod!50}\hl{yellow}), the more similar visual text is labeled as \#1 (in \sethlcolor{ForestGreen!30}\hl{green}), while the less similar one is labeled as \#2 (in \sethlcolor{Maroon!30}\hl{red}).
In the first rows, visual text is aligned to ground-truth glyphs by maximizing IoU values, with overlapped regions highlighted in red.
In the second rows, HOG features are visualized as gradient orientation fields, also highlighting overlaps in \textcolor{red}{\textbf{red}}.
Both metrics align closely with human perception, with HOG-similarity showing robust absolute values, even for thin fonts in Figure~\ref{fig:suppl_eval_metrics}(b).

\section{Font Captioning}
\label{suppl:captioning}
\begin{figure*}[t]
    \vspace{-.8em}
    \centering\small
    \includegraphics[width = .6\textwidth]{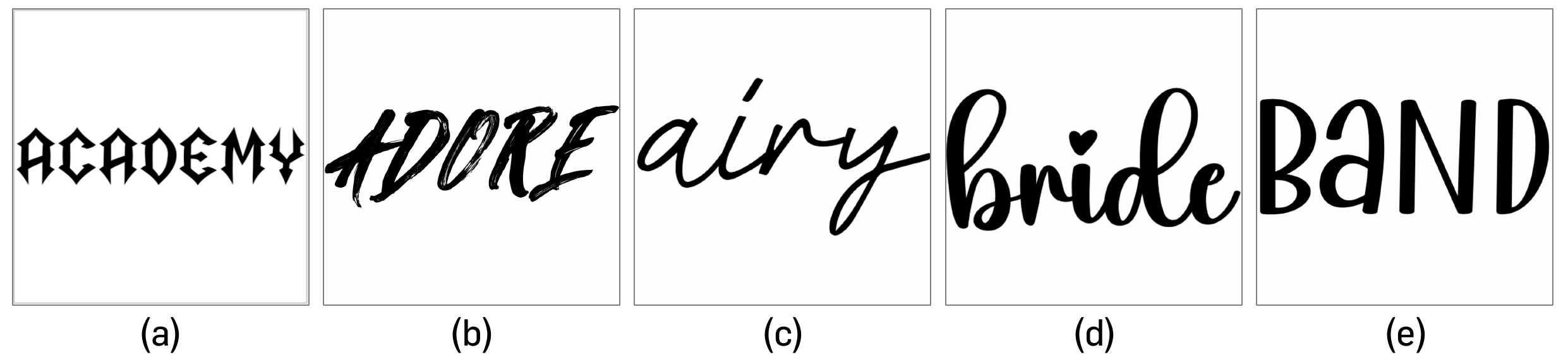}
    \vspace{-.8em}
    \caption{\textbf{Font reference examples.}}
    \label{fig:font_captions}
\end{figure*}

In Table~\ref{tab:font_captions}, we present examples of font descriptions generated by Qwen-VL~\citep{bai2023qwen}, a vision-language model, which are used for our baseline \textit{Qwen-VL captioning} (their corresponding font references are shown in Figure~\ref{fig:font_captions}).
The detailed descriptions of each font reference are generated using the query: \textit{``Question: Describe the font style of the word in detail. Answer:''}.
However, these captions are overly descriptive, including specific details of individual glyphs in the reference, which are redundant for generating other characters.
To address this, we use GPT-4o~\citep{OpenAI2024ChatGPT4o} to summarize them into concise font descriptors, retaining the key components that capture the overall font style.
The descriptors are then incorporated into the evaluation prompts to generate visual text images in the desired font styles.

\section{IPAdapter-Instruct}
\label{suppl:ipadapter_instruct}
\begin{figure}[ht]
    \centering\small
    \vspace{-0.1in}
    \includegraphics[width=\linewidth]{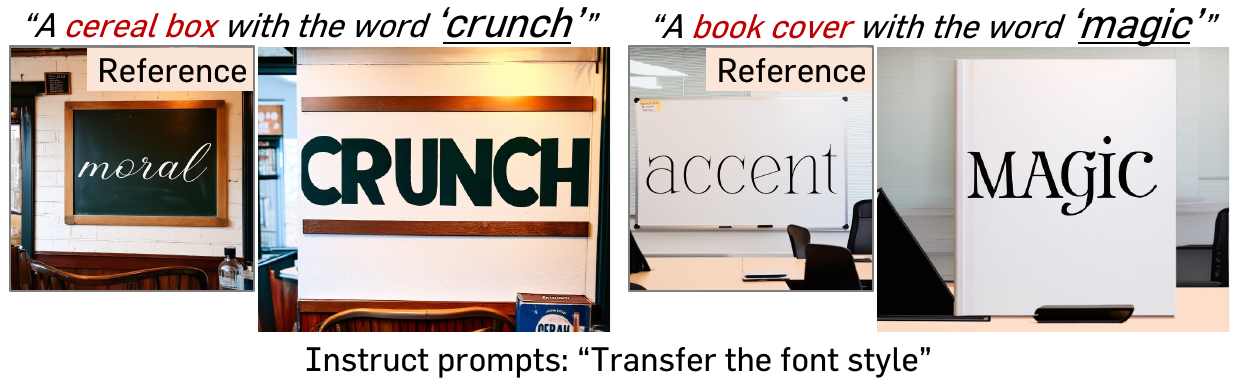}
    \vspace{-0.2in}
    \caption{\textbf{More examples from IPAdapter-Instruct.}
        The pre-trained IPAdapter-Instruct fails to effectively disentangle font styles from realistic reference images, especially when the reference glyph is embedded in real scenes.
    }
    \label{fig:suppl_ipadapter_instruct}
    \vspace{-0.15in}
\end{figure}

IPAdapter-Instruct is generally pre-trained to be conditioned on an additional instruction that specifies which aspects of the reference image should be transferred.
Hence, we use the instruction ``\textit{Transfer the font style},’’ and Figure~\ref{fig:suppl_ipadapter_instruct} shows results from IPAdapter-Instruct using conditioning images with reference glyphs.
Compared to results obtained using simple glyph images on white backgrounds (Figure~\ref{fig:teaser},~\ref{fig:qualitative}), IPAdapter-Instruct struggles to disentangle font attributes from the reference images.
Specifically, it preserves the spatial composition of the scene (\eg wooden frames on a white wall, white space above a wooden desk) rather than transferring the reference font style.

\section{User Study}
\begin{figure*}[t]
    \centering\small
    \includegraphics[width = 1.0\textwidth]{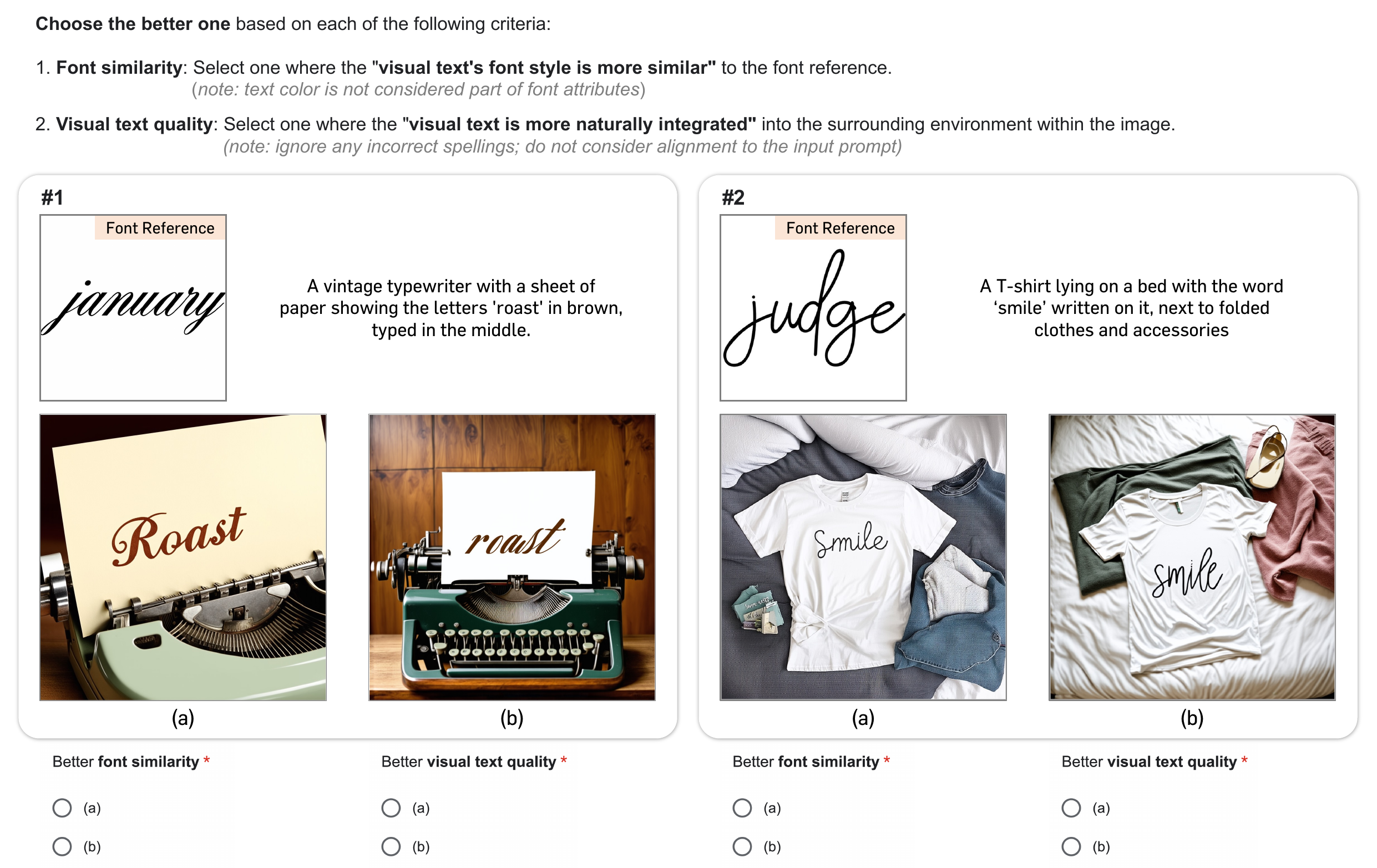}
    \caption{\textbf{User study.}
        We conduct a user study to evaluate \textit{font similarity} and \textit{visual text quality}.
        Users are presented with the global questions shown at the top of the figure and choose the better sample based on each criterion.
        To ensure consistent evaluation, each criterion is specified as follows:
        For \textit{font similarity}, the visual text’s \textit{color} is not considered part of font attributes.
        For \textit{visual text quality}, other aspects of the image, such as \textit{incorrect spellings} or \textit{prompt misalignment}, should be disregarded.
    }
    \label{fig:suppl_user_study}
\end{figure*}

We conduct a user study (Table~\ref{tab:suppl_user_study}) to evaluate two comprehensive aspects of font customization: 1) \textit{font similarity}, and 2) \textit{visual text quality} (how naturally the visual text integrates into the image context).
We sample 50 images per model, and collect 4,600 responses from 46 users.
Users are asked to select the better one between two samples, generated with the same prompt and font reference, based on each aspect (see Figure~\ref{fig:suppl_user_study}).
The study demonstrates that \methodnameshort’s outputs are strongly preferred over the baselines, highlighting its superior performance in font similarity.
This also validates the alignment of our font similarity metrics with human perception, where \textit{Qwen-VL captioning} achieves higher scores on font similarity than IPAdapter-Instruct.
For visual text quality, the study shows that \methodname effectively integrates visual text with its background, achieving results comparable to the original SD3, even when conditioned on a specific font style.

\begin{table}[ht]
\centering
\vspace{-.5em}
\caption{\textbf{User study results.}
    \methodname strongly outperforms the baselines in font similarity while achieving comparable results to the original SD3 (\textit{Qwen-VL captioning}) in visual text quality.
}
\vspace{-.5em}
\label{tab:suppl_user_study}
\resizebox{\linewidth}{!}
{
\begin{tabular}{l c c}
    \toprule
    Win rate & vs. Qwen-VL captioning & vs. IPAdapter-Instruct \\
    \midrule
    Font similarity & 91.4\% & 83.2\% \\
    Visual text quality & 50.8\% & 82.9\% \\
    \bottomrule
\end{tabular}
}
\vspace{-1em}
\end{table}

\section{More Qualitative Examples and Prompts}
\label{suppl:more_examples_prompts}

\begin{figure*}[t]
    \centering\small
    \includegraphics[width = .98\textwidth]{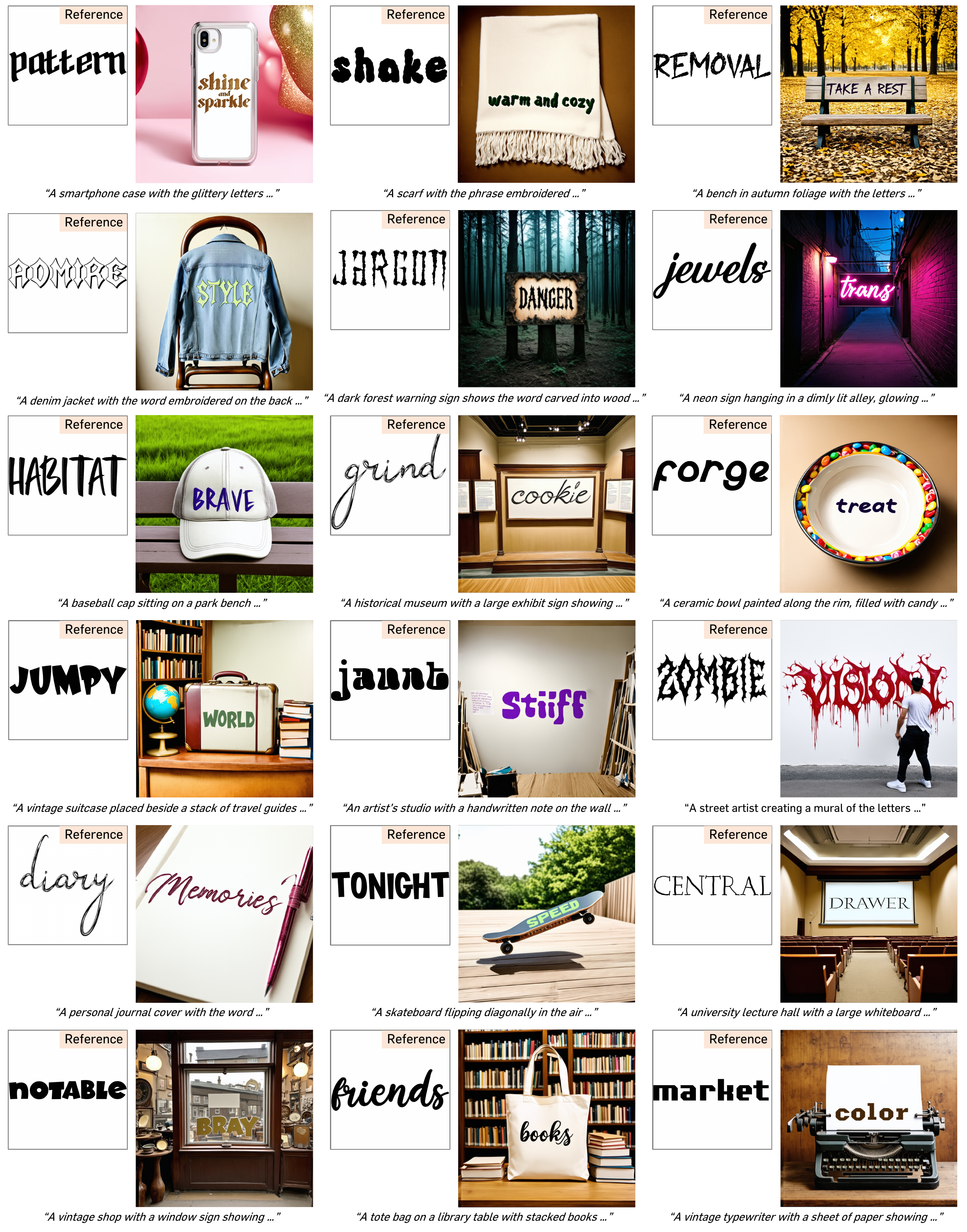}
    \caption{\textbf{More examples of visual text generation by \methodname.}}
    \label{fig:suppl_qualitative}
\end{figure*}

\begin{figure*}[t]
    \centering\small
    \includegraphics[width = \textwidth]{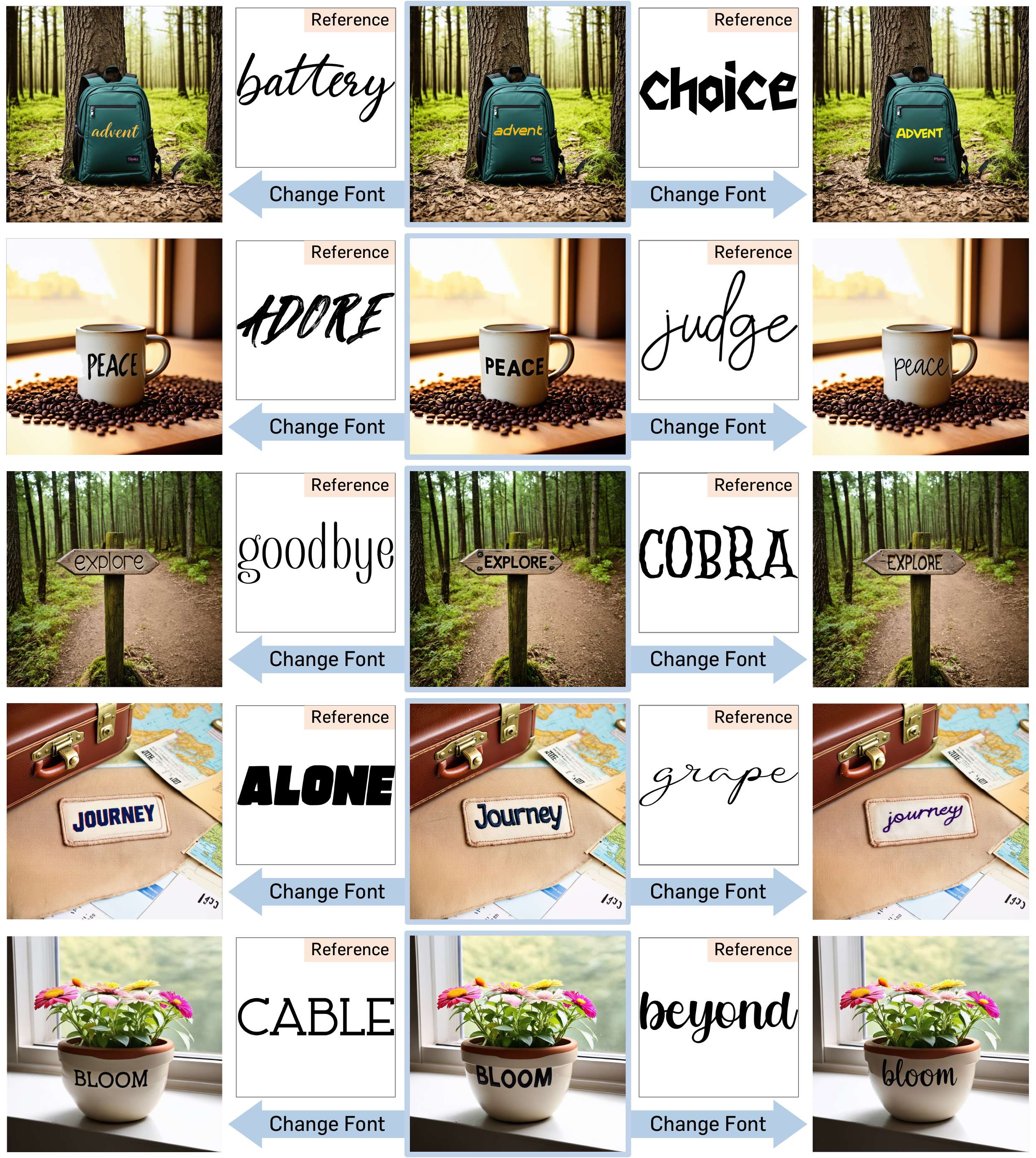}
    \caption{\textbf{More examples of visual text editing by \methodname.}}
    \label{fig:suppl_edit}
\end{figure*}

\begin{figure*}[t]
    \centering\small
    \includegraphics[width=0.7\textwidth]{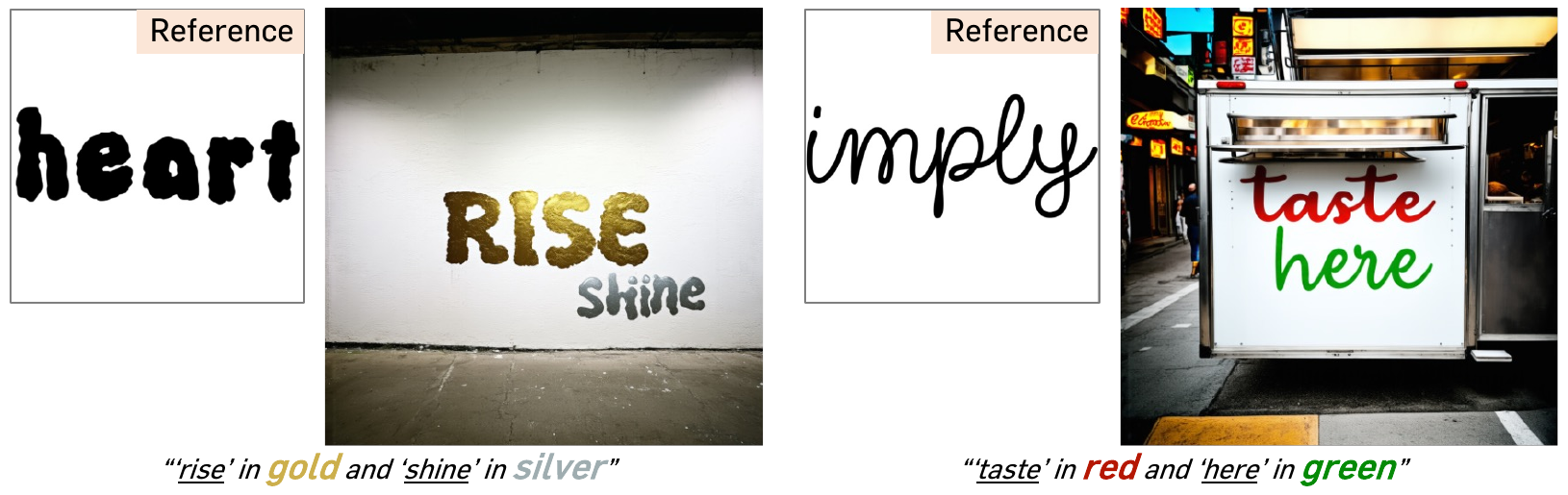}
    \caption{\textbf{Examples of multi-color text control by \methodname.}
    Text color can be easily controlled via textual prompting, without affecting the quality of font customization---demonstrating effective disentanglement between font styles and other glyph features.
    }
    \label{fig:suppl_multiple_colors}
\end{figure*}

\begin{figure*}[t]
    \centering\small
    \includegraphics[width=\textwidth]{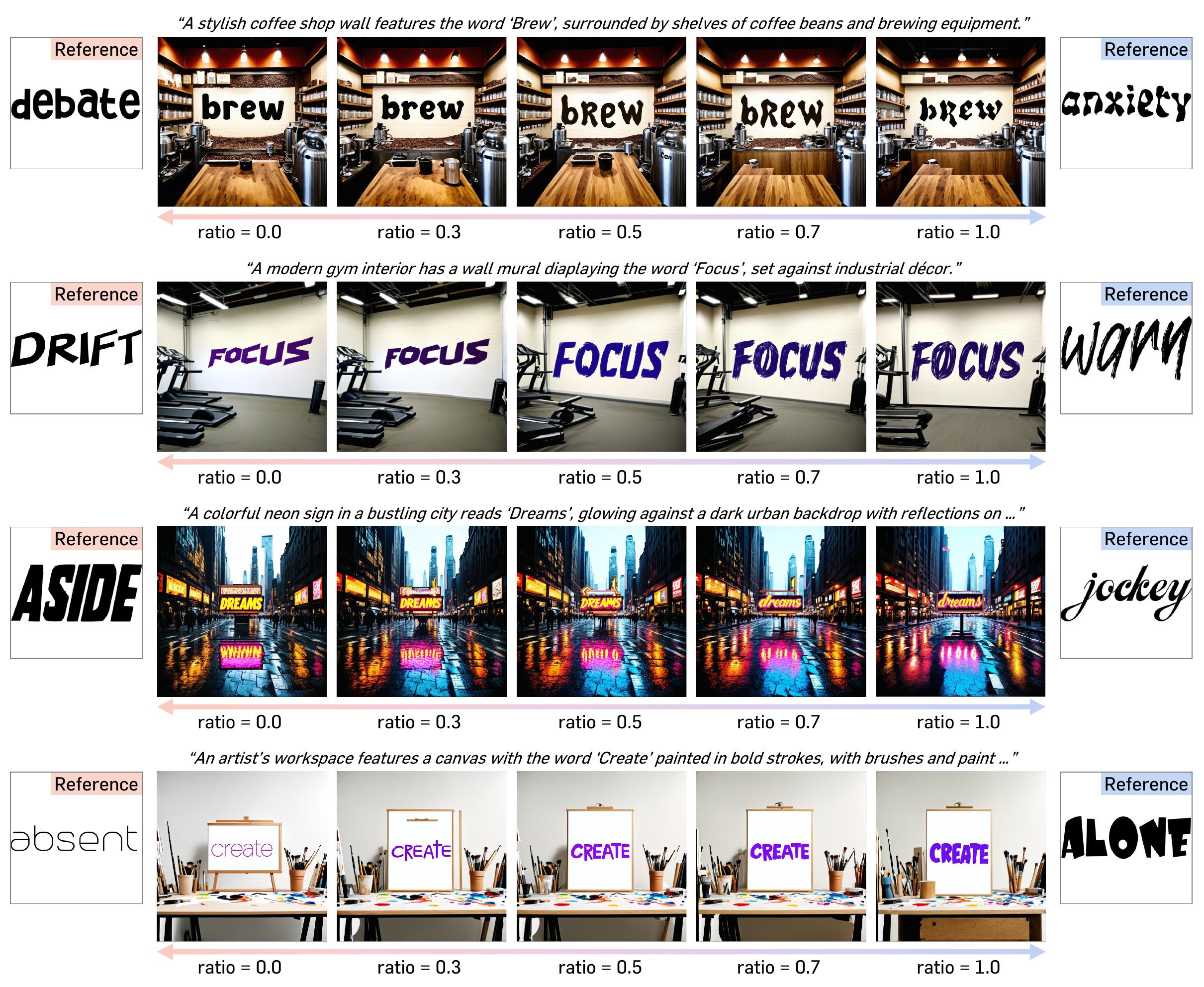}
    \caption{\textbf{More examples of font blending by \methodname.}}
    \label{fig:suppl_blend}
\end{figure*}

We present additional results from \methodname for several tasks: visual text generation in Figure~\ref{fig:suppl_qualitative}, visual text editing in Figure~\ref{fig:suppl_edit}, multi-color text control in Figure~\ref{fig:suppl_multiple_colors}, and font blending in Figure~\ref{fig:suppl_blend}.
For reference, we also include the detailed prompts used to generate the samples in Figure~\ref{fig:teaser} and Figure~\ref{fig:qualitative}, shown in Table~\ref{tab:suppl_gen_prompts}.

\section{Limitations}
Since our SD3 backbone is limited to generating text in English, our method supports multilingual font references while restricting target text generation to English. 
However, extending this approach to multilingual text generation would be straightforward with a multilingual backbone and remains an avenue for future work.

\begin{figure}[ht]
    \centering\small
    \vspace{-0.1in}
    \includegraphics[width=\linewidth]{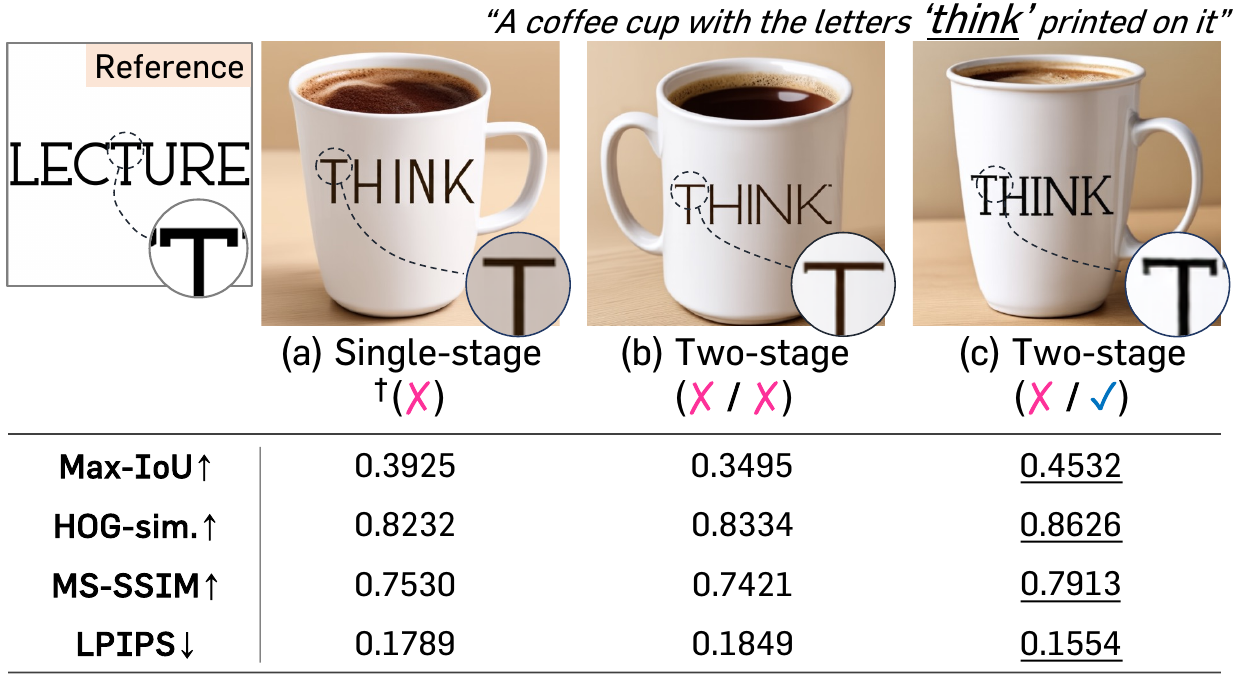}
    \vspace{-0.2in}
    \caption{\textbf{Figure~\ref{fig:freeze_resampler} with font similarity metrics.}
    While our font metrics successfully reflect overall font fidelity with respect to the reference, subtle yet meaningful differences results in relatively small metric variations.
    }
    \label{fig:suppl_freeze_resampler_with_metrics}
\end{figure}

Additionally, our proposed font metrics primarily capture overall glyph structure rather than fine-grained details.
As a result, although higher metric values correlate with improved font fidelity, subtle yet meaningful enhancements may not be fully emphasized quantitatively, as illustrated in Figure~\ref{fig:suppl_freeze_resampler_with_metrics}.
Future work could explore alternative evaluation methods, such as VLM-based metrics, to better quantify these nuanced, fine-grained differences.

\begin{table*}[t!]
\centering
\caption{\textbf{Detailed prompts for generated backgrounds.}
    Each original prompt is used to generate each background image in Figure~\ref{fig:suppl_scene_text}, created using GPT-4o to include empty regions for placing visual texts.
    For training, random glyphs (indicated as \textbf{bold} letters) are added to these backgrounds, and the original prompts are adjusted to incorporate proper visual text descriptions, with the keyword ‘\underline{empty}’ removed.
}
\label{tab:suppl_bg_prompts}
\resizebox{\linewidth}{!}
{
\begin{tabular}{l l}
    \toprule
    Position & Original prompt + \sethlcolor{Magenta!15}\hl{\textit{Visual text description}} \\
    \midrule
    Row 1, Col 1 & ``An \underline{empty} chalkboard in a rustic pub'' + \sethlcolor{Magenta!15}\hl{\textit{``, featuring the word \textbf{‘inquiry’} written on it''}} \\
    Row 1, Col 2 & ``An \underline{empty} signpost in a picturesque vineyard'' + \sethlcolor{Magenta!15}\hl{\textit{``, bearing the word \textbf{‘ABLE’} written on it''}} \\
    Row 1, Col 3 & ``An \underline{empty} billboard in a futuristic city''  + \sethlcolor{Magenta!15}\hl{\textit{``, showing the word \textbf{‘team’} prominently displayed''}} \\
    Row 1, Col 4 & ``An \underline{empty} sign at a historic church''  + \sethlcolor{Magenta!15}\hl{\textit{``, displaying the word \textbf{‘DWELL’} written on it''}} \\
    Row 1, Col 5 & ``An \underline{empty} poster frame in a trendy music venue''  + \sethlcolor{Magenta!15}\hl{\textit{``, featuring the word \textbf{‘partner’} written on it''}} \\
    Row 2, Col 1 & ``An \underline{empty} poster frame in a chic boutique''  + \sethlcolor{Magenta!15}\hl{\textit{``, displaying the word \textbf{‘fiction’} written on it''}} \\
    Row 2, Col 2 & ``An \underline{empty} billboard in a vibrant festival''  + \sethlcolor{Magenta!15}\hl{\textit{``, showing the word \textbf{‘impure’} written on it''}} \\
    Row 2, Col 3 & ``A majestic mountain with an \underline{empty} sign''  + \sethlcolor{Magenta!15}\hl{\textit{``, featuring the word \textbf{‘close’} written on it''}} \\
    Row 2, Col 4 & ``An \underline{empty} wooden sign in a serene garden pavilion''  + \sethlcolor{Magenta!15}\hl{\textit{``, bearing the word \textbf{‘THOUGHT’} written on it''}} \\
    Row 2, Col 5 & ``An \underline{empty} announcement screen at a bustling tech expo'' + \sethlcolor{Magenta!15}\hl{\textit{``, showing the word \textbf{‘ample’} clearly displayed''}} \\
    Row 3, Col 1 & ``An \underline{empty} poster frame in a chic bar''  + \sethlcolor{Magenta!15}\hl{\textit{``, presenting the word \textbf{‘WALK’} written on it''}} \\
    Row 3, Col 2 & ``An \underline{empty} board in an office''  + \sethlcolor{Magenta!15}\hl{\textit{``, displaying the word \textbf{‘FAIL’} written on it''}} \\
    Row 3, Col 3 & ``An \underline{empty} banner at a vibrant parade''  + \sethlcolor{Magenta!15}\hl{\textit{``, highlighting the word \textbf{‘BICYCLE’} written on it''}} \\
    Row 3, Col 4 & ``An \underline{empty} billboard in a scenic coastal town''  + \sethlcolor{Magenta!15}\hl{\textit{``, displaying the word \textbf{‘phrase’} on it''}} \\
    Row 3, Col 5 & ``An \underline{empty} wooden sign in a serene lakeside cabin'' + \sethlcolor{Magenta!15}\hl{\textit{``, showcasing the word \textbf{‘SCANNER’} engraved on it''}} \\
    \bottomrule
\end{tabular}
}
\end{table*}
\begin{table*}[t]
\centering
\caption{\textbf{Examples of Evaluation Prompts.}
    Each subset consists of 300 prompts, categorized by the level of image context complexity: \textsc{Simple}, \textsc{Moderate}, and \textsc{Complex}.
    The placeholder $<$*$>$ is replaced with a random word, and the color of each visual text is specified to assess controllability over text colors as a distinct text attribute.
}

\vspace{-2mm}

\label{tab:suppl_eval_prompts}
\resizebox{\linewidth}{!}
{
\begin{tabular}{l}
    \toprule
    \underline{\textsc{Simple} (\textit{object})} \\
    ``A chalkboard with the letters `$<$*$>$' in white.''\\
    ``A book cover with the letters `$<$*$>$' in blue.''\\
    ``A metal plaque engraved with the letters `$<$*$>$' in purple.''\\
    ``A graffiti wall with the letters `$<$*$>$' in navy.''\\
    $\cdots$\\

    \midrule
    \underline{\textsc{Moderate} (\textit{object + background})} \\
    ``A coffee cup with the letters `$<$*$>$' in violet, in a cafe window.''\\
    ``A T-shirt with the letters `$<$*$>$' in purple, displayed on a clothing rack.''\\
    ``A chalkboard sign with the letters `$<$*$>$' in white, outside a bakery.''\\
    ``A wall poster showing the letters `$<$*$>$' in orange, in an art gallery.''\\
    $\cdots$\\

    \midrule
    \underline{\textsc{Complex} (\textit{object + background + extra contexts})} \\
    ``A rainy street with neon signs reflecting on the pavement, one of the signs displays the letters `$<$*$>$' in blue.''~~~~~~~~~~~~~~~~~~~~~~~~~~~~~~~~~~~~~~~~~~~~~~~~\\
    ``A modern office space with a presentation screen showing the letters `$<$*$>$' in black, in a meeting.''\\
    ``A train station platform with a digital sign displaying the letters `$<$*$>$' in green, as people wait for the train.''\\
    ``A peaceful garden with a wooden sign engraved with the letters `$<$*$>$' in brown, near a fountain.''\\
    $\cdots$\\
    \bottomrule
\end{tabular}
}
\end{table*}

\begin{table*}[t]
\sethlcolor{Magenta!15}
\centering
\vspace{-1em}
\caption{\textbf{Font captioning.}
    Each row corresponds to each font reference in Figure~\ref{fig:font_captions}.
    The font references are initially described in detail by Qwen-VL~\citep{bai2023qwen} (keywords are highlighted in \hl{red}) and then summarized into concise \textit{font descriptors} using GPT-4o~\citep{OpenAI2024ChatGPT4o}.
}
\vspace{-.8em}
\label{tab:font_captions}
\footnotesize
\begin{tabular}{c p{9.8cm} p{5.2cm}}
    \toprule
    Sample & Described by QWen-VL & Summarized by GPT-4o \\
    \midrule
    (a) & The word ``ACADEMY'' in the image is set in a black \hl{gothic} font style. The letters are slightly curved and have \hl{jagged edges}, giving them a medieval or \hl{medieval-inspired} look. The font is \hl{thick} and stands out against the white background. The overall effect is that of an old-fashioned or classic style, which is fitting for a gothic Academy. & \textit{``the letters written in a gothic style, thick with jagged edges, medieval-inspired''}\\
    (b) & The word ``ADORE'' in the image is written in a \hl{brush style} font. The letters have a slightly curved shape, with each letter \hl{connected} to the one before it and the one after it, creating a smooth and \hl{flowing effect}. The letters are black and stand out against the white background. The font style gives the word an \hl{elegant and artistic} look. & \textit{``the letters written in a brush style, flowing and connected with an artistic, elegant look''} \\
    (c) & The word ``airy'' in the image is written in a \hl{cursive} font, which gives it a \hl{flowing and graceful} appearance. The letters are \hl{connected smoothly}, without any hard edges, and the lines are slightly curved, giving the word a \hl{sense of movement and elegance}. The font style is handwritten, which adds a personal and organic touch to the word. The letters ``i'' and ``y'' are particularly connected, with the ``i'' partially covering the ``y,'' which further emphasizes the flow and connection of the letters. Overall, the font style gives the word a unique and attractive appearance. & \textit{``the letters written in a flowing cursive style, connected letters with graceful, smooth lines, offering elegance and movement''} \\
    (d) & The word ``bride'' in the image is written in a \hl{cursive} font. The letters have an organic, \hl{flowing shape, with slight curves} and bends that give them a natural and \hl{graceful} appearance. The letters are \hl{connected} to each other, \hl{with a small heart-shaped symbol} between the letters ``i'' and ``e.'' The heart shape is simple and minimalistic, with two small, equal-length curved lines forming the shape. The overall font style is \hl{elegant and romantic}, suitable for use as a wedding-related term. & \textit{``the letters written in an elegant cursive style, flowing with slight curves, connected with a small heart symbol, romantic and graceful''} \\
    (e) & The word ``BAND'' in the image is written in a \hl{hand-drawn} style font. The letters have a \hl{slightly curved} shape, with the \hl{line thickness varying slightly} from one letter to another. The letters ``B'' and ``D'' are slightly wider than the others, creating a sense of balance and harmony with the rest of the word. The overall font style is \hl{casual and relaxed}, giving the word a friendly and \hl{approachable look}. & \textit{``the letters written in a casual hand-drawn style, curved and slightly varying in thickness, giving a relaxed and approachable look''} \\
    \bottomrule
\end{tabular}
\end{table*}
\begin{table*}[t]
\captionsetup{justification=justified, singlelinecheck=false}
\caption{\textbf{Specific prompts.}
    Each sample is generated with its corresponding prompt, while font references are shown in separate figures.
}
\centering

\vspace{-2mm}

\label{tab:suppl_gen_prompts}
\resizebox{\linewidth}{!}
{
\footnotesize
\begin{tabular}{l l l}
    \toprule
    Image & Position & Prompt \\
    \midrule

    Figure~\ref{fig:teaser}(a) & Row 1 & ``A backpack resting against a tree with the phrase `\textit{\textbf{explore the unknown}}' written on it, surrounded by camping gear and a map.'' \\
    & Row 2 & ``A notebook with the phrase `\textit{\textbf{write down all your great ideas}}' written on the cover, placed on a wooden desk.'' \\
    & Row 3 & ``A snow globe with the phrase `\textit{\textbf{winter memories captured in this}}', sitting on a mantelpiece.'' \\
    \midrule

    Figure~\ref{fig:teaser}(b) & Top & ``A cozy coffee shop with a mug displaying the letters `\textit{\textbf{Anise}}' in red, warm atmosphere.'' \\
    & Bottom & ``A cozy coffee shop with a mug displaying the letters `\textit{\textbf{Angel}}' in purple, warm atmosphere.'' \\
    \midrule

    Figure~\ref{fig:teaser}(c) & Left & ``A fashion brand’s hoodie design with the letters `\textit{\textbf{Freedom}}' on the back, streetwear look.'' \\
    & Right & ``A bustling market with a spice counter displaying the letters `\textit{\textbf{Spice}}', aromatic scents.'' \\
    \midrule

    Figure~\ref{fig:qualitative} & Col 1 & ``A bustling market with a large banner displaying the letters `\textit{\textbf{Silver}}' above the main entrance.'' \\
    & Col 2 & ``A lively street at night with a bright neon sign showing the letters `\textit{\textbf{Lone}}' in blue, above a restaurant.'' \\
    & Col 3 & ``A rustic lodge with a hand-carved sign displaying the letters `\textit{\textbf{Graced}}' in brown, above the entrance' \\
    & Col 4 & ``A vintage typewriter with a sheet of paper showing `\textit{\textbf{Stand}}' typed in the middle.'' \\
    & Col 5 & ``An outdoor event with a white banner displaying the letters `\textit{\textbf{Hitch}}', above the crowd.'' \\
    & Col 6 & ``A historical museum with a large exhibit sign showing the letters `\textit{\textbf{Halt}}' in black'' \\
    \midrule

    Figure~\ref{fig:app_crossling} & Row 1, Col 1 & ``A metal sign with the letters `\textit{\textbf{Legacy}}' engraved on it.'' \\
    & Row 1, Col 2 & ``An art gallery display plaque showing the word `\textit{\textbf{Heritage}}', celebrating cultural preservation.'' \\
    & Row 1, Col 1 & ``A neon sign hanging in a dimly lit alley, glowing the letters `\textit{\textbf{Light}}', against the night.'' \\
    & Row 2, Col 2 & ``A ceramic teapot on display with the word `\textit{\textbf{Serenity}}' painted on it, enhancing the peaceful setting.'' \\
    \midrule

    Figure~\ref{fig:app_edit} & Top & ``A t-shirt with the letters `\textit{\textbf{Parsley}}' written in red.'' \\
    & Bottom & ``A piece of glass with the letters `\textit{\textbf{YAH}}' etched on it in black.'' \\
    \bottomrule
\end{tabular}
}
\end{table*}

\clearpage

\end{document}